\documentclass[10pt,twocolumn,letterpaper]{article}
\pdfoutput=1

\usepackage{cvpr_modified}
\usepackage{times}
\usepackage{epsfig}
\usepackage{graphicx}
\usepackage{amsmath}
\usepackage{amssymb}

\usepackage{wrapfig}
\usepackage{color}
\usepackage{subcaption}
\usepackage{booktabs}

\usepackage[pagebackref=true,breaklinks=true,letterpaper=true,colorlinks,bookmarks=false]{hyperref}

\cvprfinalcopy 

\def\cvprPaperID{1941} 

\ifcvprfinal\fi

\DeclareMathOperator*{\argmin}{arg\,min}

\DeclareMathOperator{\x}{\mathbf{x}}

\newcommand{\Fig}{Fig.}

\newcommand*{\Scale}[2][4]{\scalebox{#1}{$#2$}}
\DeclareMathOperator*{\median}{\mathtt{median}}

\begin{document}

\title{Optical Flow in Mostly Rigid Scenes}

\author{Jonas Wulff \quad \quad Laura Sevilla-Lara \quad \quad Michael J.~Black\hspace{0.1in}\\ 
    Max Planck Institute for Intelligent Systems, T\"{u}bingen, Germany\\
    {\tt\small  \{jonas.wulff,laura.sevilla,black\}@tue.mpg.de}
       }

\maketitle
\thispagestyle{empty}


\begin{abstract}
The optical flow of natural scenes is a combination of the motion of the observer and the independent motion of objects.
Existing algorithms typically focus on either recovering motion and structure under the assumption of a purely static world or optical flow for general unconstrained scenes.
We combine these approaches in an optical flow algorithm that estimates an explicit segmentation of moving objects from appearance and physical constraints.
In static regions we take advantage of
strong constraints to jointly estimate the camera motion and the 3D structure of the scene over multiple frames. This allows us to also regularize the structure instead of the motion. Our formulation uses a Plane+Parallax framework, which works even under small baselines, and reduces the motion estimation to a one-dimensional search problem, resulting in more accurate estimation. 
In moving regions the flow is treated as unconstrained, and computed with an existing optical flow method. The resulting {\em Mostly-Rigid Flow (MR-Flow)} method achieves state-of-the-art results on both the MPI-Sintel and KITTI-2015 benchmarks.
\end{abstract}

\section{Introduction}
\noindent
The world is composed of things that move and things that do not.
The 2D motion field, which is the projection of the 3D scene motion onto the image plane, arises from observer motion relative to the static scene and the independent motion of objects.
A large body of work exists on estimating camera motion and scene structure in purely static scenes,
generally referred to as Structure-from-Motion (SfM).
On the other hand, methods that estimate general 2D image motion, or optical flow, make much weaker assumptions about the scene. 
Neither approach fully exploits the mixed structure of natural scenes.
Most of what we see in such scenes is static - houses, roads, desks, etc.\footnote{In KITTI-2015 and MPI-Sintel, independently moving regions make up only 15\% and 28\% of the pixels, respectively.}
Here, we refer to these static parts of the scene as the \textit{rigid scene}, or \textit{rigid regions}.
At the same time, moving objects like people, cars, and animals make up a small but often important part of natural scenes.
Despite the long history of both SfM and optical flow, no state-of-the art optical flow method synthesizes both into an algorithm that works on general scenes like those in the MPI-Sintel dataset \cite{Butler:ECCV:Sintel} (Fig.~\ref{fig:teaser}).
In this work, we propose such a method to estimate 
optical flow in video sequences
of generic scenes that contain moving objects within a rigid scene.

For the rigid scene, the camera motion and depth structure fully determine the motion, which forms the basis of SfM methods.
Modern optical flow benchmarks, however, are full of moving objects such as cars or bicycles in KITTI, or humans and dragons in Sintel.
Assuming a fully static scene or treating these moving objects as outliers is hence not viable for optical flow algorithms; we want to reconstruct flow everywhere.

Independent motion in a scene typically arises from well defined objects with the ability to move.  This points to a possible solution.
Recently, convolutional neural networks  (CNN) have achieved good performance on detecting and segmenting objects in images, and have been successfully incorporated into optical flow methods \cite{Bai:2016:SemanticDeepFlow,Sevilla:SOF}.
Here we take a slightly different approach.
We modify a common CNN and train it on novel data to obtain a rigidity score from the labels, taking into account that some objects (\eg humans) are more likely to move than others (\eg houses).
This score is combined with additional motion cues to obtain an estimate of rigid and independently moving regions.

After partitioning the scene into rigid and moving regions, we can deal with each appropriately.
Since the motion of moving objects can be almost arbitrary, it is best computed using a classical unconstrained flow method.
The flow of the rigid scene, on the other hand, is extremely restricted, and only depends on the depth structure and the camera motion and calibration.
In theory, one could use an existing SfM algorithm to reconstruct the camera motion and the 3D structure of the scene, and project this structure back to obtain the motion of the rigid scene regions. Two factors make this hard in practice.
First, the number of frames usually considered in optical flow is small; most methods only work on two or three consecutive frames.
SfM algorithms, on the other hand, require tens or hundreds of frames to work reliably.
Second, SfM algorithms require large camera baselines in order to reliably estimate the fundamental matrices.
In video sequences, large baselines are rare, since the camera usually translates very little between frames.
An exception to this are automotive scenarios such as the KITTI benchmark, where the recording car often moves rapidly and the frame rate is low.

Since full SfM is unreliable in general flow scenarios,
we adopt the \textit{Plane+Parallax} (P+P) framework~\cite{Irani:2002:MultiFramePPP,Irani:1998:ReferenceFrames,Sawhney:1994:3DGeometryPPP}
In this framework, frames are registered to a common plane, which is aligned in all images after the registration.
This removes the motion caused by camera rotation and simple intrinsic camera parameter changes, leaving parallax
as the sole source of motion.
Since all parallax is oriented towards or away from a common focus of expansion in the frame, computing the parallax is reduced to a 1D search problem and therefore easier than computing the full optical flow.

Here we show that using the P+P framework brings an additional advantage: the parallax can be factored into a structure component, which is independent of the camera motion and constant across time, and a temporally varying camera component, which is a \emph{single number per frame}.
We integrate the structure information across time; by definition, the structure of the rigid scene does not change.
By combining the structure information from multiple frames, our algorithm generates a better structure component for all frames, and fills in areas that are unmatched in a single pair of frames due to occlusion.

Additionally, the relationship between the structure component and the parallax (and thus, the optical flow) enables us to regularize the flow in a physically meaningful way, since regularizing the structure implicitly regularizes the flow.
We use a robust second-order regularizer, which corresponds to a locally planar prior.

\begin{figure*}
\includegraphics[width=\textwidth]{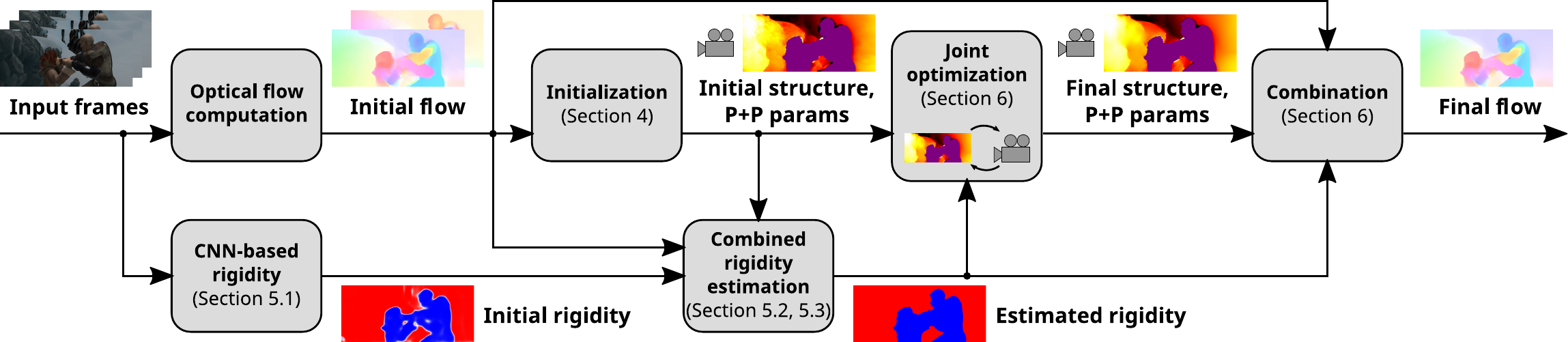}
\caption{Algorithm overview.
Given a triplet of frames, we first compute initial flow and an initial rigidity estimate based on a semantic segmentation CNN.
The images are then aligned to a common plane, and the initial flow is converted to an estimate of the structure in the rigid scene using the Plane+Parallax framework.
Where the P+P constraints are violated, the rigidity is refined, while at the same time the structure is refined using a variational optimization.
To obtain the final flow estimate, the initial flow is used in moving regions, while the refined structure induces the flow in the rigid scene.}
\label{fig:pipeline}
\end{figure*}


We integrate the regularization into a novel objective function measuring the photometric error across three frames as a function of the structure and camera motion.
This allows us to optimize the structure and also to recover from poor initializations.
We call the method MR-Flow for {\em Mostly-Rigid Flow} and show an overview in Fig.~\ref{fig:pipeline}.


We test MR-Flow on MPI-Sintel~\cite{Butler:ECCV:Sintel} and KITTI 2015~\cite{Menze:2015:OSF}
(Fig.~\ref{fig:teaser}).
Among published monocular methods, at time of writing, we achieve the lowest error on MPI-Sintel on both passes;
on KITTI-2015, our accuracy is second only to~\cite{Bai:2016:SemanticDeepFlow}, a method specifically designed for automotive scenarios.
Our code, the trained CNN, and all data is available at~\cite{MRFlow:Website}.

In summary, we present three main contributions.
First, we show how to segment the scene into rigid regions and independently moving objects, allowing us to estimate the motion of each type of region appropriately.
Second, we extend previous plane+parallax methods to express the flow in the rigid regions via its depth structure.
This allows us to regularize this structure instead of the flow field and to combine information across more than two frames.
Third, we formulate the motion of the rigid regions as a single model.
This allows us to iterate between estimating the structure and to recover from unstable initializations.

\section{Previous work}
\noindent
SfM and optical flow have both made significant, but mostly independent, progress.
Roughly speaking, SfM methods require purely rigid scenes and use sparse point matches, wide
baselines between frames, solve for accurate camera intrinsics
and extrinsics, and exploit bundle adjustment to optimize over many
views at once. 
In contrast, optical flow is applied to scenes containing generic motion,
exploits continuous optimization, makes weak
assumptions about the scene (\eg that it is piecewise smooth),
and typically processes only pairs of video frames at a time.

{\bf Combining optical flow and SfM.}
There have been many attempts to combine SfM and flow methods,
dating to the 80's \cite{HeegerJepson:1992}.
For video sequences from narrow-focal-length
lenses, the estimation of the camera motion is challenging, 
as it is easy to confuse translation with rotation
and difficult to estimate the camera intrinsics~\cite{direct}.

More recently there have been attempts to combine SfM and
optical flow~\cite{Bai:2016:SemanticDeepFlow,Oisel:1998:SPIE,Valgaerts:2009:DAGM,Wedel:2009:ICCV,Yamaguchi:2013:Epipolarflow}.
The top monocular optical flow method on the KITTI-2012
benchmark estimates the fundamental matrix and computes flow along the
epipolar lines \cite{Yamaguchi:2013:Epipolarflow}.
This approach is limited to fully rigid scenes.
Wedel \etal~\cite{Wedel:2009:ICCV} compute the fundamental matrix and
regularize optical flow to lie along the epipolar lines.
If they detect independent motion, they revert to standard optical flow for the entire frame.
In contrast, we segment static from moving {\em regions} and use appropriate constraints within each type of region.
Roussos \etal~\cite{Roussos:2012:ISMAR} assume a known calibrated camera and solve for depth, motion and segmentation of a scene with moving objects.  They perform batch
processing on sequences of about 30 frames in length, making this more
akin to SfM methods.  While they have impressive results, they consider relatively
simple scenes and do not evaluate flow accuracy on
standard benchmarks.

{\bf Plane+Parallax.}
P+P methods were developed in the mid-90's
\cite{Irani:2002:MultiFramePPP,Sawhney:1994:3DGeometryPPP}.
The main idea is that stabilizing two frames with a planar motion
(homography) removes the camera rotation and simplifies the
geometric reasoning about structure \cite{Irani:PAMI:1997,Triggs:2000:ECCV}.
In the stabilized pair, motion is always oriented towards or away from the epipole
and corresponds to parallax, which is related to the distance of the point
from the plane in the 3D scene.

Estimating a planar homography can be done robustly and with more stability
than estimating the fundamental matrix
\cite{Irani:1998:ReferenceFrames,Irani:PAMI:1997}. 
While one is not able to estimate metric depth, the planar
stabilization simplifies the matching process, turning the 2D optical
flow estimation problem into a 1D problem that is equivalent to stereo
estimation.
Given the practical benefits, one may ask why P+P
methods are not more prevalent in the leader boards of
optical flow benchmarks. 
The problem is that such methods work only for rigid scenes.
Making the P+P approach usable in general natural scenes is one of our main contributions.

{\bf Moving region segmentation.}
There have been several attempts to segment moving scenes into regions
corresponding to independently moving objects by exploiting 3D motion
cues and epipolar motion \cite{Adiv:1985:PAMI,Thompson:1990:IJCV,Weber:PAMI:1997}.
Several methods use the P+P framework to detect independent motions, but those
methods typically only do detection and not flow estimation, and are
often applied to simple scenes where there is a dominant motion
like the ground plane and small moving objects
\cite{Irani:1998:MovingObjectDetection,Sawhney:2000:PAMI,Yuan:2007:PAMI}.
Irani \etal \cite{Irani96efficientrepresentations} develop mosaic representations that
include independently moving objects but do not explicitly compute their
flow. 
Given two frames as input, Ranftl \etal~\cite{Ranftl:2016:DenseMonocularDepthEstimation} segment a general moving scene into piecewise-rigid components and reason about the depth and occlusion relationships.
While they produce impressive depth estimates,
they rely on accurate flow estimates between the frames and do not refine the flow itself.




{\bf Combining multiple flow methods.}
There is also existing work on combining motion
estimates from different algorithms into a single estimate \cite{FusionFlow,Brostow},
but these do not attempt to fuse rigid and general motion.
Bergen \etal \cite{Bergen:ECCV:1992} define a
framework for describing optical flow problems using different
constraints from rigid motion to generic flow, but do not combine
these models into a single method. 



Recent work combines segmentation and flow. Sevilla \etal \cite{Sevilla:SOF} perform semantic segmentation
and use different models for different semantic classes.
Unlike them, we use semantic segmentation to estimate the rigid scene and then impose stronger geometric constraints in
these regions.
Hur and Roth~\cite{Hur2016} integrate semantic segmentation over time, leading to more accurate flow estimation for objects and better segmentation performance.


Most similar to our approach is~\cite{Bai:2016:SemanticDeepFlow}, which first segments the scene into objects using a CNN.
A fundamental matrix is then computed and used to constrain the flow within each object.
Our work is different in a number of important ways.
(i) Their approach is sequential and cannot recover from an incorrect fundamental matrix estimate.
We propose a unified objective function where the parts of the solution inform and improve each other.
(ii) \cite{Bai:2016:SemanticDeepFlow} relies exclusively on the CNN to segment moving regions.
While this works in specific scenarios such as automotive, it may not
generalize to new scenes. 
We combine semantic segmentation and motion to classify
rigid regions and thus require less accurate semantic rigidity estimates.
This makes our algorithm both more robust and more general, as demonstrated by the fact that in contrast to~\cite{Bai:2016:SemanticDeepFlow} we evaluate on the challenging MPI-Sintel benchmark.
(iii) \cite{Bai:2016:SemanticDeepFlow} requires moving objects to be
rigid (\ie, rigidly moving vehicles) and assumes a small rotational
component of the egomotion.  
This works for KITTI-2015 but does not apply to more general scenes.
(iv) \cite{Bai:2016:SemanticDeepFlow} uses only two frames at a time and extrapolates into occlusions.
Our model combines information across time, and thus it is able to compute accurate flow in occlusions.









\section{Plane + Parallax background}
\noindent
The P+P paradigm has been used in rigid scene analysis for a long time.
Since it forms the foundation of our algorithm, we briefly review the parts that are important for this work and refer the reader to \cite{Irani:1998:ReferenceFrames,Sawhney:1994:3DGeometryPPP} for more details.

The core idea of P+P is to align two or more images to a common plane $\mathbf{\Pi}$, so that
\begin{equation}
\mathbf{x} = \left\langle H \mathbf{x'}_h \right\rangle \quad \forall (\mathbf{x},\mathbf{x'})
\text{ on }
\mathbf{\Pi}
\end{equation}
where $\mathbf{x}$ and $\mathbf{x'}$ represent a point in the reference frame and the corresponding point in another frame of the sequence, $\x_h$ denotes $\x$ in homogeneous coordinates, $H$ is the homography mapping the image of $\mathbf{\Pi}$ between frames, and $\left\langle \mathbf{a} \right\rangle = \left( a_1 / a_3, a_2 / a_3 \right)$ is the perspective normalization.

This alignment removes the effects of camera rotation and the effect of camera calibration change (such as a zoom) between the pair of frames~\cite{Zelnik-Manor:2000:PlanarMotion}.
Getting rid of rotation is especially convenient, since the ambiguity between rotation and translation in case of small displacements is a major source of numerical instabilities in the estimation of the structure of the scene.

When computing optical flow between aligned images, the flow of the pixels corresponding to points on the plane is zero\footnote{Note that the plane does not have to correspond to a physical surface, but merely to a rigid, ``virtual'' plane.}.
For an image point $\mathbf{x}$ corresponding to a 3D point $\mathbf{X}$ off the plane, the residual motion is given as~\cite{Sawhney:1994:3DGeometryPPP}
\begin{equation}
\label{eq:foe}
\mathbf{u_p} \left( \mathbf{x} \right)
=
\frac
{1}
{1 - \frac{d(C_2)}{T_z} \frac{z}{d(\mathbf{X})}}
\left( \mathbf{e} - \mathbf{x} \right), 
\end{equation}
where $d(C_2)$ is the distance of the second camera center to $\mathbf{\Pi}$, $z$ is the distance of point $\mathbf{X}$ to the first camera, $T_z$ is the depth displacement of the second camera, $d(\mathbf{X})$ is the distance from point $\mathbf{X}$ to $\mathbf{\Pi}$, and $\mathbf{e}$ is the common focus of expansion that coincides with the epipole corresponding to the second camera.
This representation has two main advantages.
First, instead of an arbitrary 2D vector, each flow is confined to a line; therefore computing the optical flow is reduced to a 1D search problem.
Second, when considering the flow of a pixel to different frames $t$ which are registered to the same plane,
Eq.~\eqref{eq:foe} can be written as
\begin{equation}
\label{eq:foeredux}
\mathbf{u_p} \left( \mathbf{x}, t \right)
=
\frac
{A(x) b_t}
{A(x) b_t - 1}
\left( \mathbf{e}_t - \mathbf{x} \right),
\end{equation}
where $A(x) = d(\mathbf{X})/z$ is the structural component of the flow field, which is \textit{independent of $t$}.
It is hence convenient to accumulate structure over time via $A$.
$b_t = T_z / d(C_2)$, on the other hand, encodes the camera motion to frame $t$, and is a single number per frame.
To simplify notation, we express the residual flow in terms of the \textit{parallax field} $w(\mathbf{x}, t)$, so that
\begin{equation}
\label{eq:parallax}
\mathbf{u_p} \left( \mathbf{x} \right)
=
w \left( \mathbf{x}, t \right) \frac{ \mathbf{q} }{ \Vert \mathbf{q} \Vert }, \quad
w \left( \mathbf{x}, t \right) =
\frac{ A(\mathbf{x}) b_t \Vert \mathbf{q} \Vert }
{A(\mathbf{x}) b_t - 1}, 
\end{equation}
with $\mathbf{q} = \left( \mathbf{e} - \mathbf{x} \right)$. Here, $w$ denotes the flow in pixels along the line towards $\mathbf{e}$.

We can thus parametrize the motion across multiple frames as a common structure component $A$ and per-frame parameters $\theta_t = \left\lbrace H_t, b_t, \mathbf{e}_t \right\rbrace$.
Since we use the center frame of a triplet of frames as the reference and compute the motion to the two adjacent frames, from here on we denote the two parameter sets as $\theta^+ = \left\lbrace H^+, b^+, \mathbf{e}^+ \right\rbrace$ for the forward direction and $\theta^-$ for the backward direction.

\section{Initialization}
\label{sec:initialization_alignment}
\noindent
Given a triplet of images
and a coarse, image-based rigidity estimation (described in Sec.~\ref{sec:rigidity_cnn}), the goal of our algorithm is to compute (i) a segmentation into rigid regions and moving objects and (ii) optical flow for the full frame.
We start by computing initial motion estimates using an existing optical flow method~\cite{Menze2015GCPR}.
For a triplet of images $\lbrace I^-, I, I^+ \rbrace$, we compute four initial flow fields, $\mathbf{u_0^+}$ from $I$ to $I^+$ and $\mathbf{u_0^-}$ from $I$ to $I^-$, and their respective backwards flows $\mathbf{\bar{u}_0^+}$ and $\mathbf{\bar{u}_0^-}$. 
Due to the non-convex nature of our model (see Sec.~\ref{sec:model}) we need to compute good initial estimates for the P+P parameters $\hat{\theta}^+, \hat{\theta}^-$, visibility maps $V^+, V^-$ denoting which pixels are visible and which are occluded in forward and backward directions, and an initial structure estimate $\hat{A}$.

{\bf Initial alignment and epipole detection.}
\noindent
First we compute the planar alignments (homographies) between frames.
Since P+P only holds in the rigid scene, in this section we only consider points that are marked as rigid by the initial semantic rigidity estimation.
While computing a homography between two frames is usually easy, two factors make it challenging in our case: (i) when aligning multiple frames, the plane to which the frames are aligned has to be equivalent for each frame for P+P to work, and (ii) the 3D points corresponding to the four points used to estimate the homographies have to be coplanar for Eq.~\eqref{eq:foeredux} to hold.

To compute homographies obeying these constraints, we use a two-stage process.
First, we compute initial homographies $\tilde{H}^+, \tilde{H}^-$ using RANSAC.
In each iteration, the \textit{same} random sample is used to fit both $\tilde{H}^+, \tilde{H}^-$, and a point is considered an inlier only when its reprojection error is low in both forward \textit{and} backward directions.
This ensures that the computed homographies belong to the same plane.
If a computed homography displaces the images corners by more than half the image size, it is considered invalid.
If no valid homography is found, our method returns the initial flow field. This happens on average in 2\% of the frames.


The second step is to ensure the coplanarity of the points inducing the homographies.
For this, we can turn around Eq.~\eqref{eq:foeredux}, and simultaneously refine the homographies and estimate the epipoles $\mathbf{e}^{ \lbrace +,- \rbrace }$ so that Eq.~\eqref{eq:foeredux} holds.
Let $\mathbf{u_r} = \left\langle H (\x+\mathbf{u_0})_h \right\rangle - \x$ be the residual flow after registration with $H$.
Each pair $\mathbf{x}, \mathbf{u_r}$ defines a residual flow line, and in the noise-free case, the epipole $\mathbf{e}$ is simply the intersection of these lines.
Since the computed optical flow contains noise, we compute the epipole using the method described in~\cite{MacLean:1999:RemovalOfTranslationBias}, which we found to be sufficiently robust to noise.
Therefore, $\mathbf{e}$ is a function of the optical flow and of the computed homography.
Enforcing coplanarity of the homographies is now equivalent to
enforcing that the residual flow lines in both directions each pass through a common point as well as possible.
The refined homographies are thus computed as
\begin{align}
\hat{H}^+, \hat{H}^- &= \argmin_{H^+, H^-} \sum_{\mathbf{x}} \sum_{z \in \lbrace +,- \rbrace} \rho \left( o^z (\mathbf{x}) \right),
\label{eq:homography_refinement}
\end{align}
with $o^{z}(\x)$ defining the orthogonal distance of the residual flow line at $\x$
to $\mathbf{e}^z$.
While Eq.~\eqref{eq:homography_refinement} is highly non-linear, we found that initializing with $\tilde{H}^{ \{ +,- \} }$ and using a standard non-linear minimization package such as L-BFGS~\cite{Nocedal:1980:LBFGS} produced results that greatly improved the final flow error compared to using the unrefined homographies $\tilde{H}^{ \{ +,- \} }$.
Throughout the paper, we use the Lorentzian $\rho(x) = \sigma^2 \log \left( 1 + {x^2}/{\sigma^2} \right)$ as the robust function, and compute the scaling parameter $\sigma$ via the MAD~\cite{Black:1999:EdgesAsOutliers}.
The initial epipolar estimates $\mathbf{\hat{e}}^{\lbrace +,- \rbrace }$ are computed using $\hat{H}^{ \{+,- \} }$.

To initialize $b^+,b^-$, we first compute the parallax fields by projecting $\mathbf{u}_r$ onto the parallax flow lines,
\begin{equation}
w = \mathbf{u}_r^{\top} \mathbf{q} / \Vert \mathbf{q} \Vert .
\label{eq:w_from_flow}
\end{equation}
Inserting \eqref{eq:w_from_flow} into \eqref{eq:parallax} and solving for $A$, we get
\begin{equation}
A = w / \left( b \left(\Vert \mathbf{q} \Vert - w \right)\right).
\label{eq:A_initial}
\end{equation}
Note that Eq.~\eqref{eq:foeredux} contains a scale ambiguity between the structure $A$ and the camera motion parameter $b$.
Therefore, we can freely choose one of $b^+, b^-$, which only affects the scaling of $A$; we choose $\hat{b}^+$ so that the initial forward structure $A^+$ defined by Eq.~\eqref{eq:A_initial} has a MAD of 1.
Since $A^-$ is a function of $b^-$ and should be as close as possible to $A^+$, we obtain the estimate $\hat{b}^-$ by solving
\begin{equation}
\hat{b}^- = \argmin_{b^-} \sum_{\mathbf{x}} \rho \left( \hat{A}^+(\mathbf{x}) - A^-(\mathbf{x}) \right).
\end{equation}
Using $\hat{b}^-$, we compute the initial backward structure $\hat{A}^-$ using Eq.~\eqref{eq:A_initial}, and set the full sets of P+P parameters to $\theta^+ = \{ \hat{H}^+, \hat{b}^+, \mathbf{\hat{e}}^+ \}$, and $\theta^-$ accordingly.

{\bf Occlusion estimation.}
\noindent
Pixels can become occluded in both directions. In occluded regions, we expect the flow to be wrong, since it can at best be extrapolated.
Given the initial flow fields,
we compute the visibility masks $V^+(\mathbf{x})$, $V^-(\mathbf{x})$ using a forward-backward check~\cite{Kalal2010}.

{\bf Initial structure estimation.}
\noindent
Using the computed structure maps $\hat{A}^{ \{+,- \} }$ and visibility maps $V^{ \{ +, - \} }$, the initial estimate for the full structure is 
\begin{equation}
\hat{A}(\mathbf{x}) = \frac{1}{\max(1, V^+(\mathbf{x}) + V^-(\mathbf{x}))}
\sum_{z \in \left\lbrace +, - \right\rbrace}
V^z(\mathbf{x}) A^z(\mathbf{x}).
\end{equation}

\section{Rigidity estimation}
\noindent
Different cues provide different, complementary information about the rigidity of a region.
The semantic category of an object tells us whether it is capable of independent motion, rigid scene parts have to obey the parallax constraint~\eqref{eq:foeredux}, and the 3D structure of rigid parts cannot change over time.
We integrate all of them in a probabilistic framework to estimate a rigidity map of the scene, marking each pixel as belonging to the rigid scene or to a moving object. 

\subsection{Semantic rigidity estimation}
\label{sec:rigidity_cnn}
\noindent
\noindent
We leverage the recent progress of CNNs for semantic segmentation to predict rigid and independently moving regions in the scene. In short, we model the relationship between an object's appearance and its ability to move. 

Obviously object appearance alone does not fully determine whether something is moving independently. 
A car may be moving, if driving, or static, if parked. 
However, for the purpose of motion estimation, not all errors are the same. 
Assuming an object is static when in reality it is not imposes false constraints that hurt the estimation of the global motion,
while assuming a rigid region is independently moving does little harm.
Thus, when in doubt, we predict a region to be independently moving.

The main optical flow benchmarks, KITTI-2015 and MPI-Sintel, provide different training data. While the essence of our model is the same for both, our training process varies to adapt to the available data. In both cases we start with the DeepLab architecture~\cite{ChenPKMY14}, pre-trained on the 21 classes of Pascal VOC~\cite{pascalVOC}, substitute all fully connected layers with convolutional layers,
and densify the predictions~\cite{Sevilla:SOF}. Both networks produce a rigidity score between 0 and 1 which we call the semantic rigidity probability $p_{s}$.

{\bf MPI-Sintel} contains many objects that are not contained in Pascal VOC, such as dragons. Thus using the CNN to predict a semantic segmentation is not possible. Also, no ground truth semantic segmentation is provided, so training a CNN to recognize these categories is not possible. However, the dataset provides ground truth camera calibration, depth and optical flow for the training set. With these we estimate rigidity maps that we take as ground truth. We do this by computing a fully rigid motion field, using the depth and camera calibration, and comparing it with the ground truth flow field. 
Pixels are classified as independently moving if these two fields differ by more than a small amount.
We make this data publicly available \cite{MRFlow:Website}.

We modify the last layer of the CNN to predict 2 classes, rigid and independently moving, instead of the original 21. We train using the last 30 frames of each sequence in the training set, and validate on the first 5 frames of each sequence. Sequences shorter than 50 frames are only included in the validation set. At test time, the probability of being rigid is computed at each pixel and then thresholded. Examples of the estimated rigidity maps can be seen in~\Fig~\ref{fig:rigidity}. 

In {\bf KITTI 2015}, 
some independently moving objects (\eg people) are masked out from the depth and flow ground truth.
Therefore, the approach we followed for MPI-Sintel cannot be used.
The objects in KITTI, however, appear in standard datasets like the enriched Pascal VOC. 
We modify the last layer of the network to predict the 22 classes that may be present in KITTI (\eg person or road) similar to~\cite{Sevilla:SOF}. 
We then classify an object as moving if it has the ability to move independently (\eg cars, or buses) and as rigid otherwise. 
Training details appear in the Sup.~Mat.~\cite{MRFlow:Website}.

Note that the same approach we use for KITTI can be used for general video sequences by using a generic pre-trained semantic segmentation network together with a definition of which semantic classes can move and which are static.
This allows our method to directly benefit from advances in semantic segmentation and novel, fine-grained semantic segmentation datasets.

\subsection{Physical rigidity estimation}

\newcommand{\rigidwidth}{0.2\textwidth}
\begin{figure*}[t]
\centerline{
	\includegraphics[width=\rigidwidth]{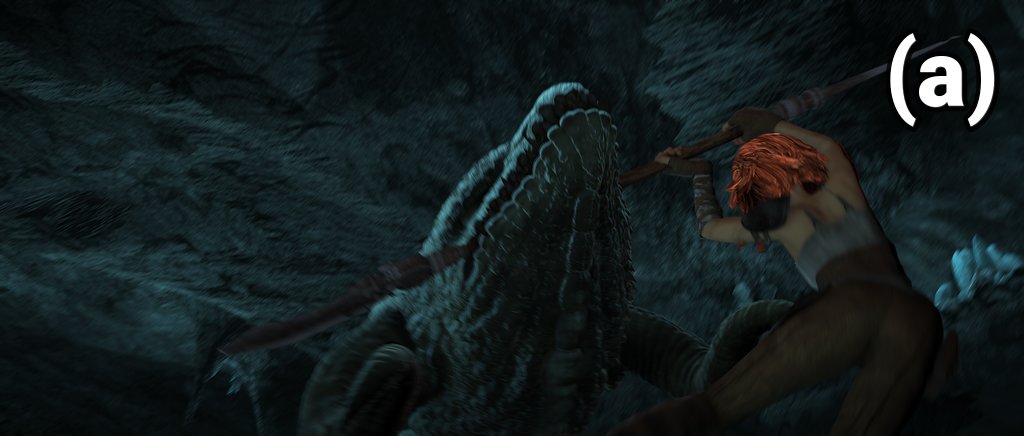}%
	\includegraphics[width=\rigidwidth]{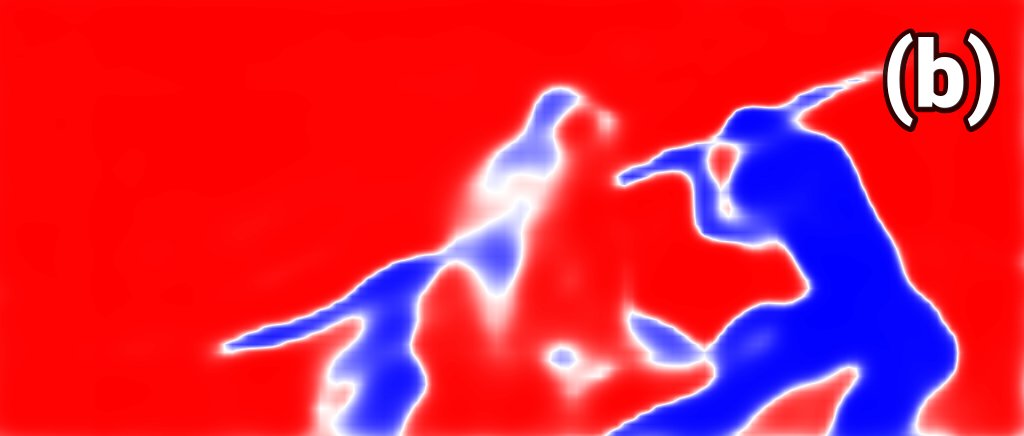}%
	\includegraphics[width=\rigidwidth]{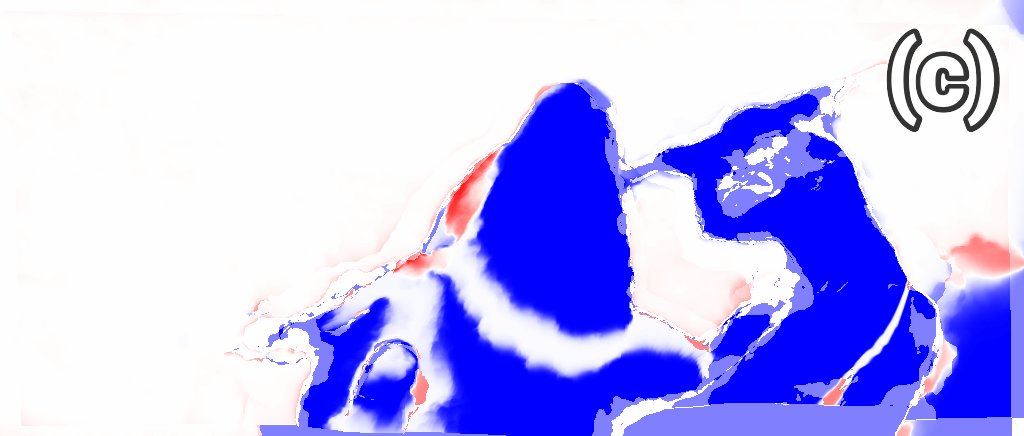}%
	\includegraphics[width=\rigidwidth]{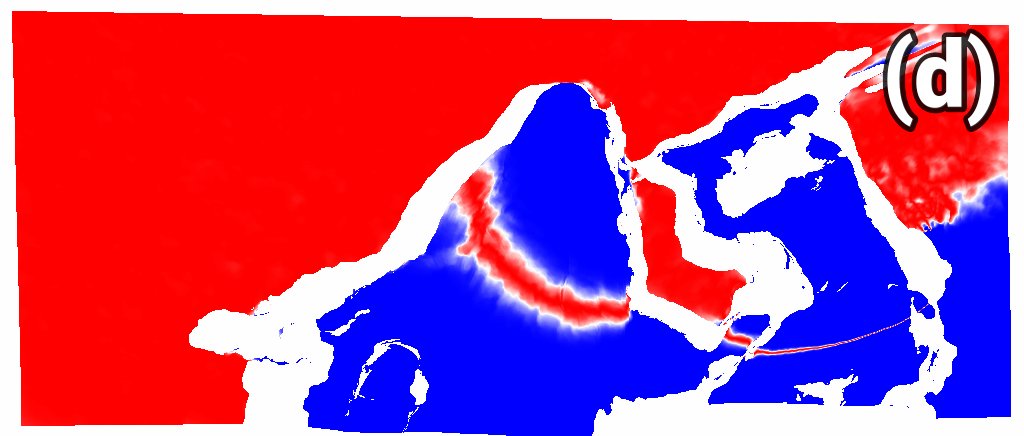}%
	\includegraphics[width=\rigidwidth]{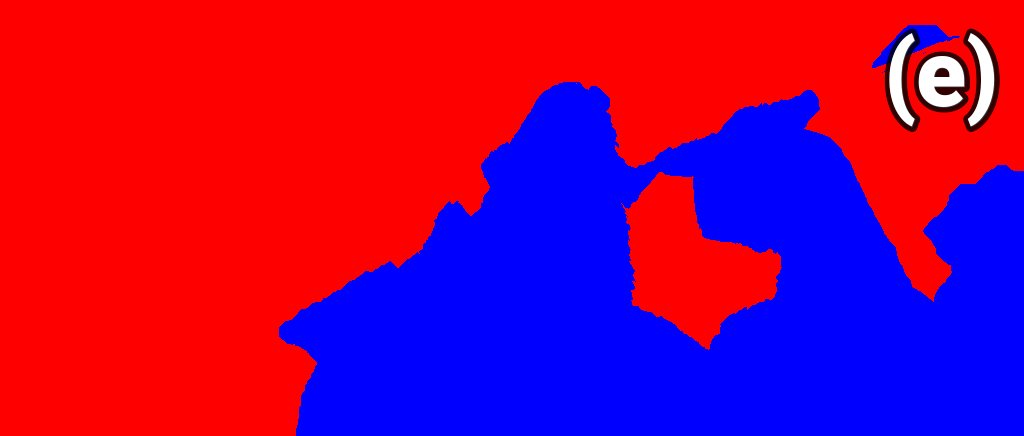}%
	}
\centerline{
	\includegraphics[width=\rigidwidth]{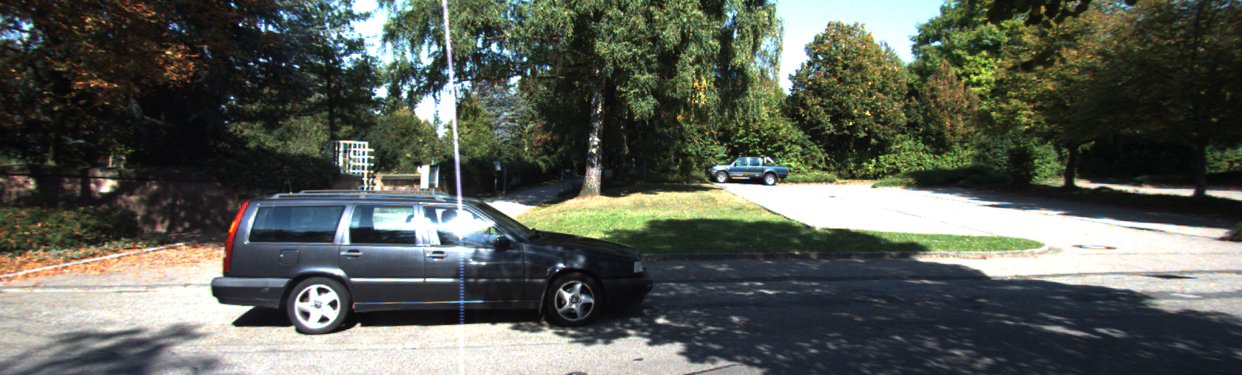}%
	\includegraphics[width=\rigidwidth]{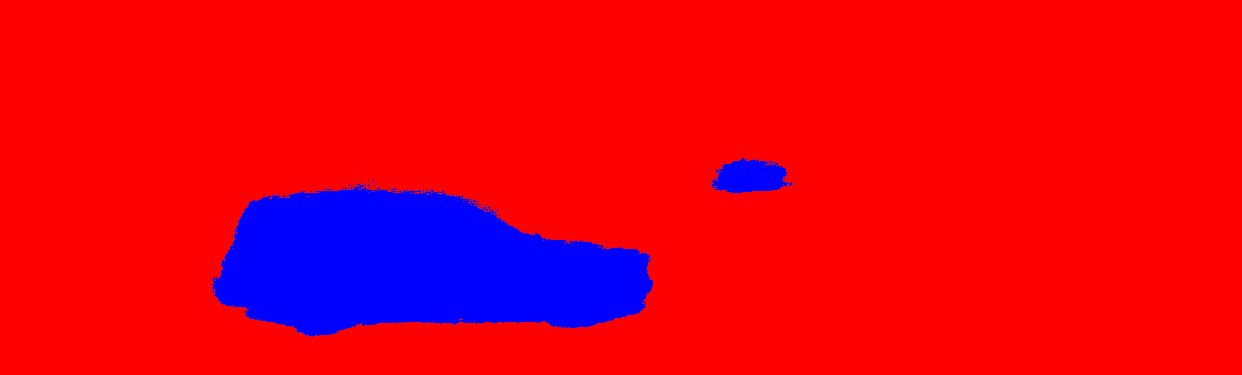}%
	\includegraphics[width=\rigidwidth]{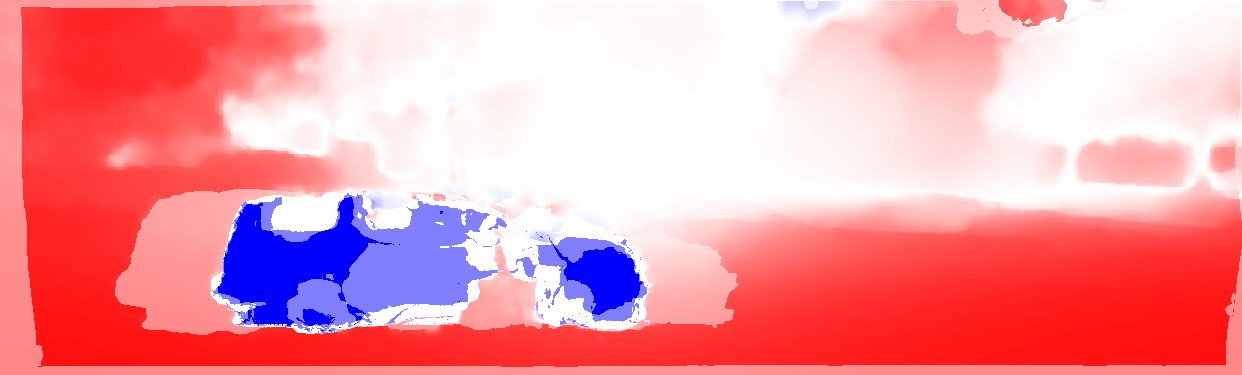}%
	\includegraphics[width=\rigidwidth]{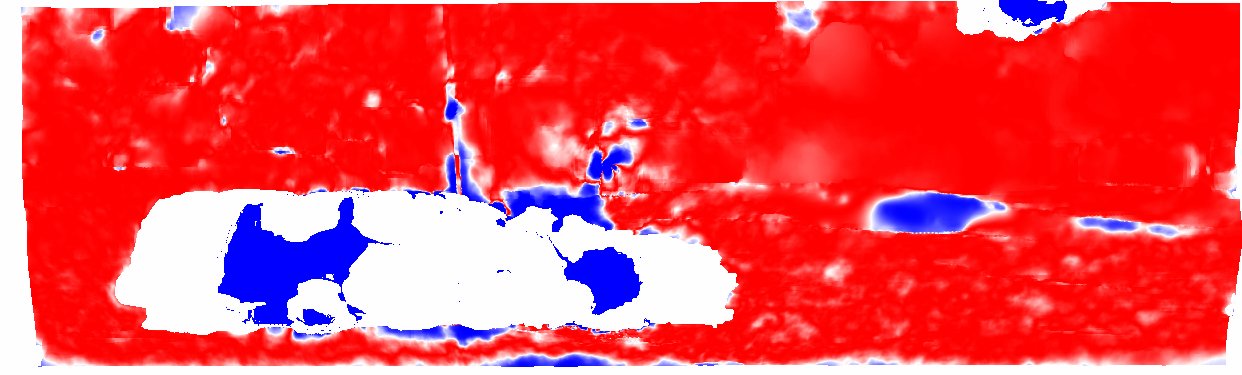}%
	\includegraphics[width=\rigidwidth]{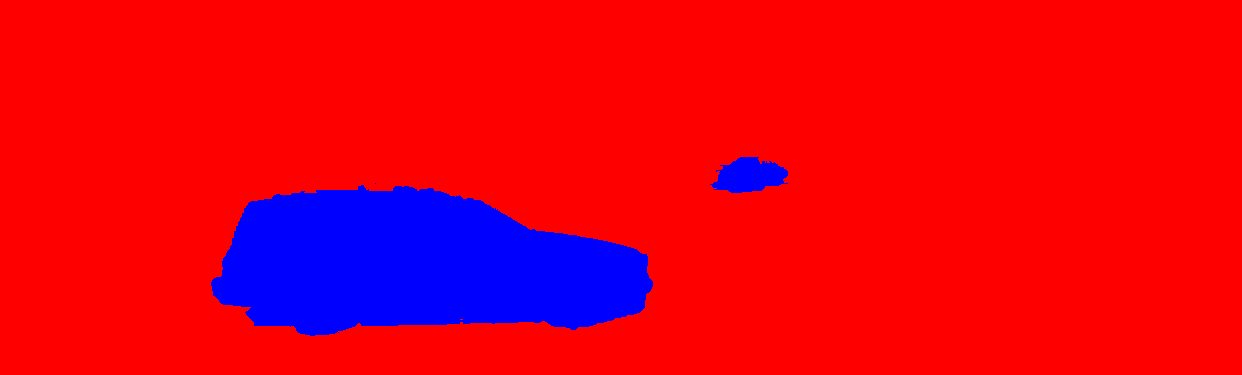}%
	}
\vspace{-0.07in}
\caption{Results of rigidity estimation on the test sets of MPI-Sintel and KITTI-2015. From an image (a), we estimate a semantic rigidity (b) and combine it with the direction-based rigidity (c) and the structure-based rigidity (d) to obtain the final estimate (e). Likely rigid regions are red, likely moving regions are blue.}
\label{fig:rigidity}
\end{figure*}
\noindent
For objects that have not been seen previously or that exhibit
phenomena like motion blur, the semantic rigidity may be wrong. 
Hence, we use two additional cues, motion direction and temporal consistency of the structure.

\noindent
{\bf Moving regions from motion direction.}
A simple approach to classify a pixel as rigid or independently moving is to test whether its parallax flow points to the epipole~\cite{Irani:1998:MovingObjectDetection}.
Here, we employ a probabilistic framework for this classification.
Due to space limitations, we just present the final result here; for the derivation, please see the Sup.~Mat.~\cite{MRFlow:Website}.


%
%
For a given point $\mathbf{x}$, our model assumes the measured corresponding point $\mathbf{x}' = \mathbf{x} + \mathbf{u_r}$ to have a Gaussian error distribution around the true correspondence with covariance matrix $\Sigma = \sigma_d^2 \mathbf{I}$.
Let $c=\Vert \mathbf{u_r} \Vert$ and $\alpha$ be the angle between $\mathbf{u_r}$ and the line connecting $\x$ to $\mathbf{e}$.
Assuming a uniform distribution of motion directions for moving objects, the likelihood of a point being rigid is then given as
\begin{equation}
p\left(\x \text{ is rigid}\right)
=
\frac{ \exp \left( - 2 t \sin^2(\alpha) \right) }
{
\exp\left( -t \right) 
	\mathbb{I}_0 \left(t\right)
+ \exp \left( - 2 t \sin^2(\alpha) \right)
}
\end{equation}
with $t = c^2 / (4 \sigma_d^2 )$ and
$\mathbb{I}_0(x)$ the modified Bessel function of the first kind.
Solving for both forward and backward directions yields the direction-based rigidity probabilities $p_{d}^+$ and $p_{d}^-$.
These are then combined into the final direction-based rigidity probability using the visibility maps
\begin{equation}
p_d =
\begin{cases}
\frac{1}{V^+ + V^-} \sum_{z\in{+,-}} V^z p_d^z
 &\text{if } V^- + V^+ > 0 \\
1/2 &\text{otherwise.}
\end{cases}
\end{equation}

\noindent
{\bf Moving regions from structure consistency.}
Another cue for rigidity is the temporal consistency of the structure.
This is particularly helpful where semantics and motion direction cannot disambiguate the rigidity, for example when an object such as a car moves parallel to the observer's motion. 

Recall that according to the P+P framework the structure of the rigid scene is independent of time.
In rigid regions that are visible in all frames, we assume the forward and backward structure $A^+$ and $A^-$ to be close to each other.
A structure based rigidity estimate $p_s$ can thus be computed as
\begin{equation}
p_{s} =
\begin{cases}
\exp \left( - \left( A^+ - A^- \right)^2 / \sigma_s^2 \right) &\text{if } V^- V^+ = 1 \\
1/2 &\text{otherwise.}
\end{cases}
\end{equation}

\noindent
{\bf Combined rigidity probability from motion.}
The motion-based probabilities $p_d$, $p_s$ can be seen as orthogonal. Surfaces that move independently along the parallax direction are considered to be rigid according to $p_d$, while surfaces that move by small amounts orthogonal to the parallax direction are considered to be rigid according to $p_s$.
Hence, for a region to be considered \textit{actually} rigid, we require both $p_d$ and $p_s$ to be high.
The final motion-based rigidity probability $p_m$ is
\begin{equation}
p_{m} =
\begin{cases}
	p_d p_s & \text{if } V^+ V^- = 1 \\
	(p_d + p_s) / 2 &\text{otherwise.}
\end{cases}
\end{equation}

\subsection{Combining rigidity estimates}
\noindent
The previously computed rigidity probabilities $p_{c}$, $p_{m}$
yield per-pixel rigidity probabilities.
To combine those into a coherent estimate, we first compute a rigidity unary
\begin{equation}
p_r = \lambda_{r,c} p_c + \left( 1- \lambda_{r,c} \right) p_m
\label{eq:rigidity_unaries}
\end{equation}
and the corresponding energy
\begin{equation}
E_r(R, \mathbf{x}) =
\begin{cases}
	-\log p_{r}(\mathbf{x}) &\text{if } R(\mathbf{x}) = 1 \\
	-\log \left( 1 - p_{r}(\mathbf{x}) \right) &\text{otherwise, }
\end{cases}
\end{equation}
with $R(\x) = 1$ if $\x$ is rigid, and 0 otherwise.
Since we expect the rigidity to be spatially coherent, 
we estimate the full labelling by solving $\hat{R} =$
\begin{equation}
\argmin_{R}
\sum_\mathbf{x}
E_r \left( R, \mathbf{x} \right)
+ \lambda_{r,p} \sum_{y\in \mathcal{N}(\mathbf{x})} w_{\mathbf{x},\mathbf{y}} \left[ R(\mathbf{x}) \neq R(\mathbf{y}) \right]
\label{eq:labelling}
\end{equation}
where $w_{\mathbf{x},\mathbf{y}}$ is the image-based Potts modulation from~\cite{Rother:Grabcut} and $\mathcal{N}(\mathbf{x})$ is the 8-connected neighborhood of $\mathbf{x}$.
Eq.~\eqref{eq:labelling} is solved using TRWS~\cite{TRWS}.

Figure~\ref{fig:rigidity} (top) shows the importance of combining different cues to recover from errors and accurately estimate the rigidity.
The semantic estimation (b) misses a large part of the dragon's head, while both the direction-based (b) and structure-based estimations misclassify different segments of the scene.
Combining cues yields a good estimate (e).
\section{Model and optimization}
\label{sec:model}
\noindent
{\bf Model.}
The final structure should fulfill a number of criteria.
First, as in the classical flow approach, warping the images using the flow induced by the structure should result in a low photometric error.
Second, we assume that our initial flow fields are reasonable, hence, the final structure should be similar to the structures defined by the initial forward and backward flow.
Third, the structure directly corresponds to the surface structure of the world, and thus we can regularize it using a locally planar model. This implicitly regularizes the flow in a more geometrically meaningful way than traditional priors on the flow.

Under these considerations, the full model for the motion of the rigid parts of the scene is defined as $E(A,\theta^+,\theta^-) =$
\begin{align}
\sum_{\x} \hat{R}(\x) \left( E_{d} + \lambda_c E_{c} + \lambda_{1st} E_{1st} + \lambda_{2nd} E_{2nd} \right).
\label{eq:energy_full}
\end{align}
$E_{d}$ is the photometric error, modulated by the estimated visibilities in forward and backward directions:
\begin{align}
E_{d} =& 
V^+(\x) \rho\left( I_a^+\left( s \left(\x, A, \theta^+ \right) \right) - I_a(\x) \right)
	\nonumber \\
&+ V^-(\x) \rho\left( I_a^-\left( s \left(\x, A, \theta^- \right) \right) - I_a(\x) \right),
\end{align}
where $I_a^-,I_a,I_a^+$ are augmented versions of $I^-,I,I^+$, \ie stacked images containing the respective grayscale images and the gradients in $x$ and $y$ directions.
The warping function $s(\x, A, \theta)$ defines the correspondence of x according to the structure $A$ and the P+P parameters $\theta$,
\begin{equation}
s(\x, A, \theta) = \left\langle H^{-1} \left( \x + \frac{A(\x)b}{A(\x)b-1}\left( \mathbf{e}-\x \right) \right)_h \right\rangle.
\end{equation}
The consistency term $E_{c}$ encourages similarity between $A$ and $A^{ \{ +,- \} }$.
\begin{align}
E_{c} = 
V^+ \rho_c \left( A - A^+ \right)
+V^- \rho_c \left( A - A^- \right) .
\end{align}
To ensure a constant error for all $A \in \left[ A^-, A^+ \right]$, we use the Charbonnier function as the robust penalty $\rho_c$.

The locally-planar regularization uses a 2nd order prior,
\begin{align}
E_{2nd} &= 
		w_x \rho \left( \nabla_{xx} A \right) \nonumber \\
&+
		w_x w_y \rho \left( \nabla_{xy} A \right)
		+
		w_y \rho \left( \nabla_{yy} A \right).
\end{align}
Here, $w_x, w_y$ are again the modulation terms from~\cite{Rother:Grabcut}, and, using a slight abuse of notation, $\nabla_{xx}, \nabla_{xy}, \nabla_{yy}$ are the second derivative operators.
Since the second order prior by itself is highly sensitive to noise, we add a first order prior
\begin{align}
E_{1st} = w_x \rho \left( \nabla_x A \right)
		+
		w_y \rho \left( \nabla_y A \right),
\end{align}
where $\nabla_x, \nabla_y$ are the first derivative operators in the
horizontal and vertical direction respectively.

{\bf Optimization.}
To minimize the energy~\eqref{eq:energy_full} we employ an iterative scheme, and alternate between optimizing for $A$ with $\theta^{ \{+,-\} }$ fixed, and for $\theta^{ \{+,-\} }$ with $A$ fixed.
When optimizing $A$, we use a standard warping-based variational optimization~\cite{Brox:TheoryOfWarping} with 1 inner and 5 outer iterations and no downscaling.
To optimize for $\theta$, we first optimize for $H, b$ using L-BFGS and then recompute $\mathbf{e}$ as described in Sec.~\ref{sec:initialization_alignment}.
We use two iterations, since we found that more do not decrease the error significantly.
This yields the final estimates $\bar{A}, \bar{\theta}^+, \bar{\theta}^-$ for the structure and the P+P parameters.

Due to the non-convex nature of~\eqref{eq:energy_full}, a global optimum is not guaranteed.
However, in practice we found that our initializations are close to a good optimum, and hence our optimization procedure works well.

{\bf Final flow estimation.}
Finally, we convert the estimated structure $\bar{A}$ into an optical flow field
\begin{equation}
\mathbf{u_s}(\mathbf{x}) = s \left(\mathbf{x}, \bar{A}, \bar{\theta}^+ \right) - \mathbf{x}.
\end{equation}
In the moving regions, we use the initial forward flow $\mathbf{u_0^+}$, and compose the full flow field as
\begin{equation}
\mathbf{u} \left( \x \right) = \hat{R}(\x) \mathbf{u_s} + \left( 1 - \hat{R}(\x) \right) \mathbf{u_0^+}.
\end{equation}

\section{Experiments}
\newcommand{\resultwidth}{0.20\textwidth}
\begin{figure*}[ht]
\centerline{
		\includegraphics[width=\resultwidth]{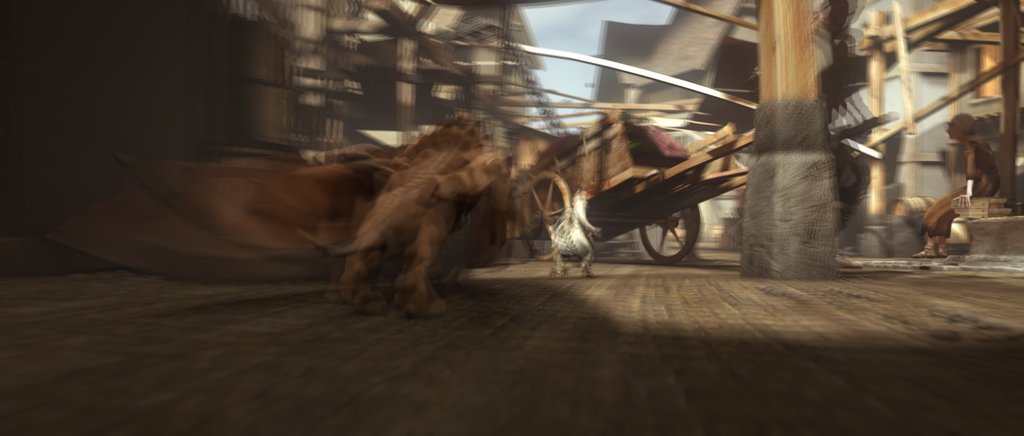}%
		\includegraphics[width=\resultwidth]{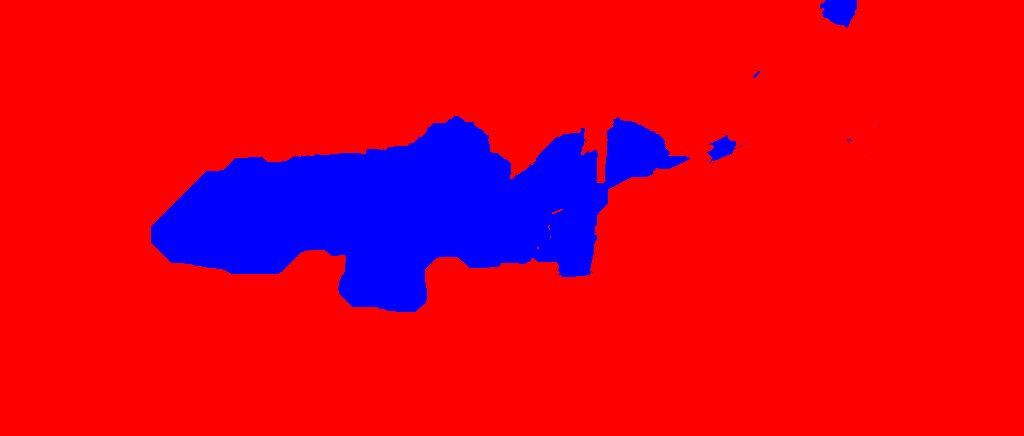}%
		\includegraphics[width=\resultwidth]{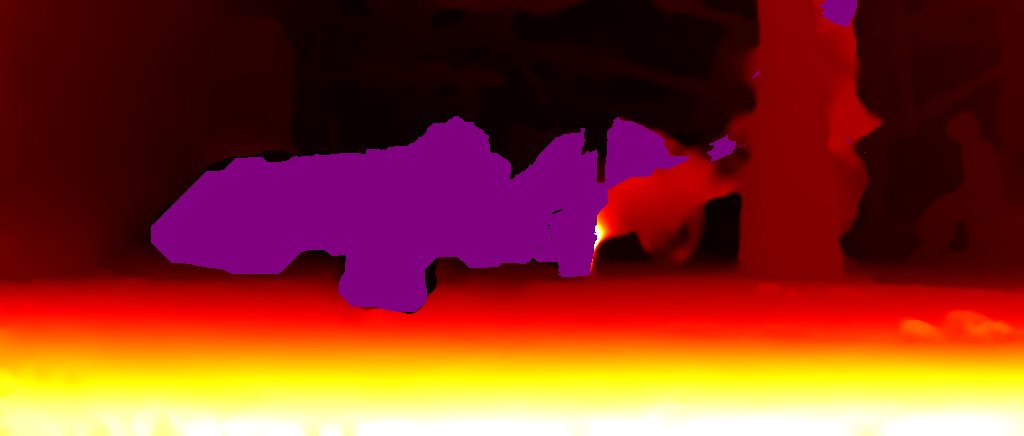}%
		\includegraphics[width=\resultwidth]{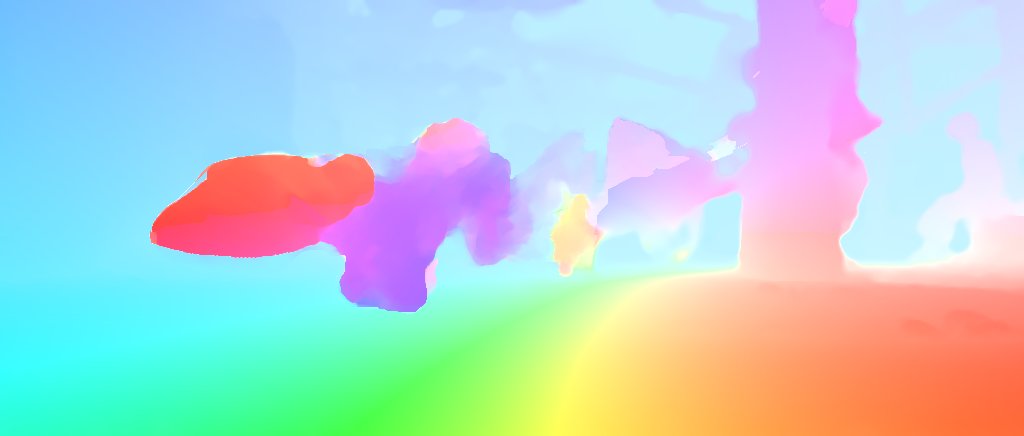}%
		\includegraphics[width=\resultwidth]{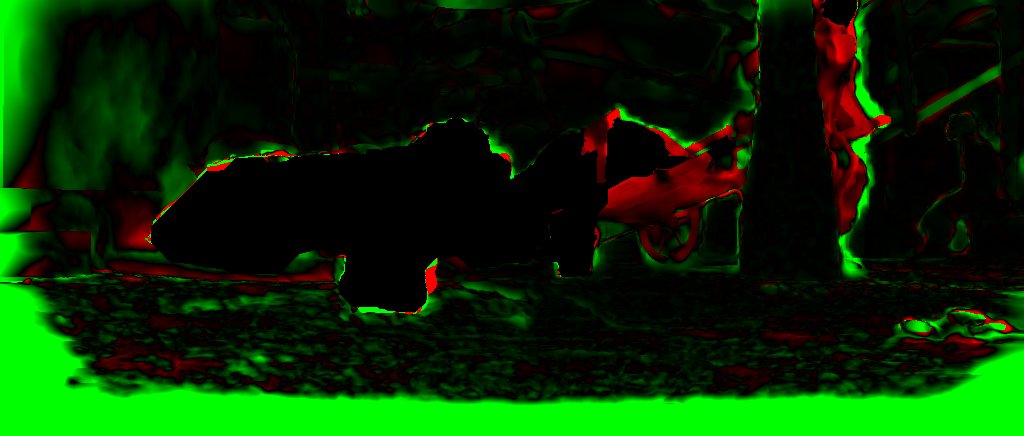}%
		}%
\centerline{
		\includegraphics[width=\resultwidth]{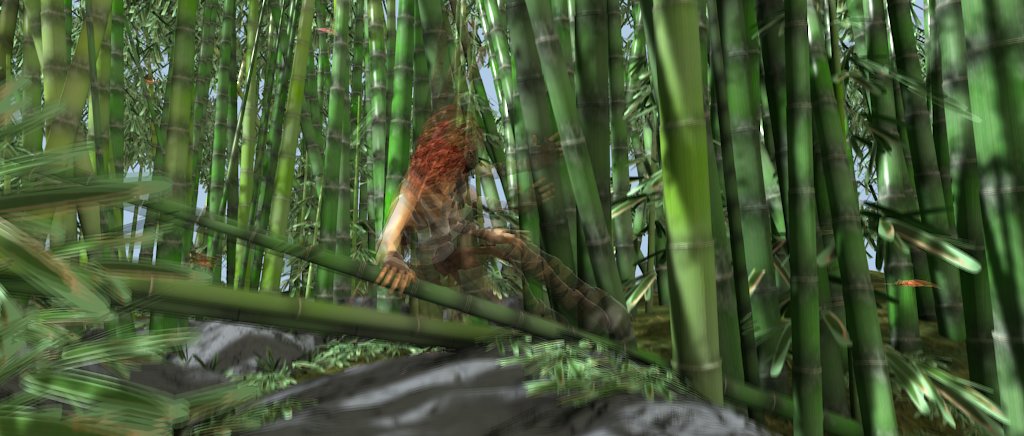}%
		\includegraphics[width=\resultwidth]{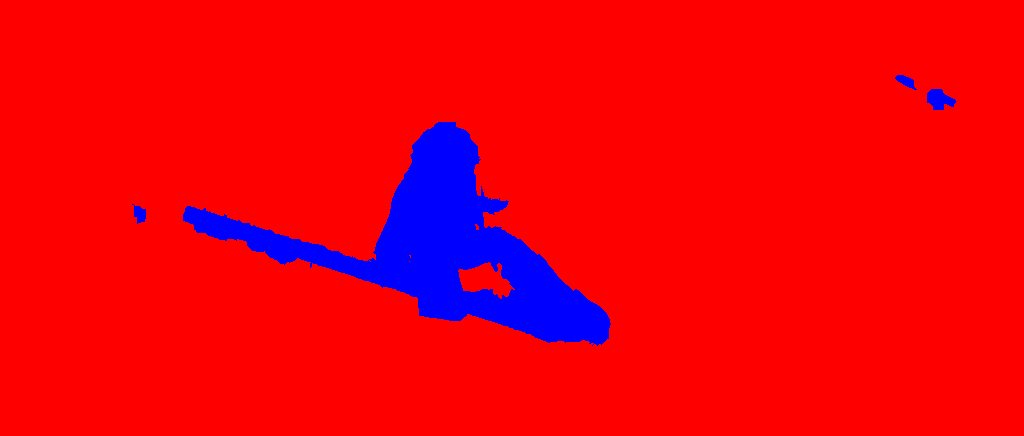}%
		\includegraphics[width=\resultwidth]{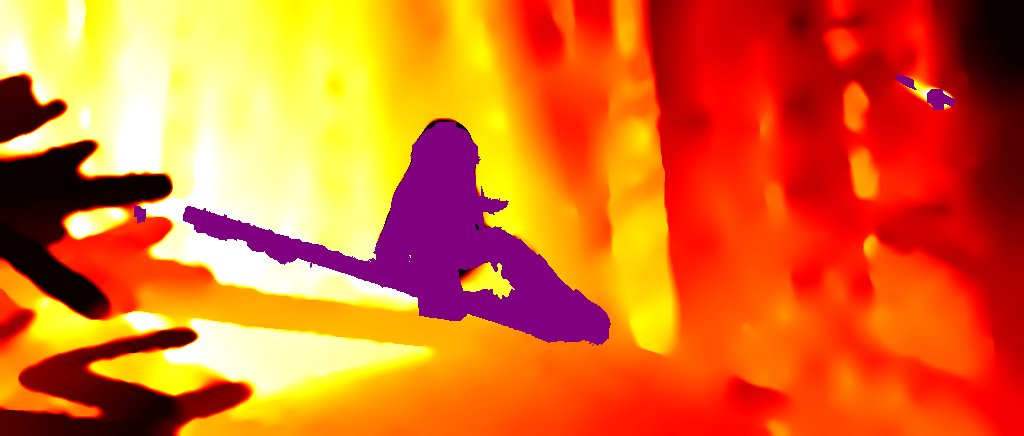}%
		\includegraphics[width=\resultwidth]{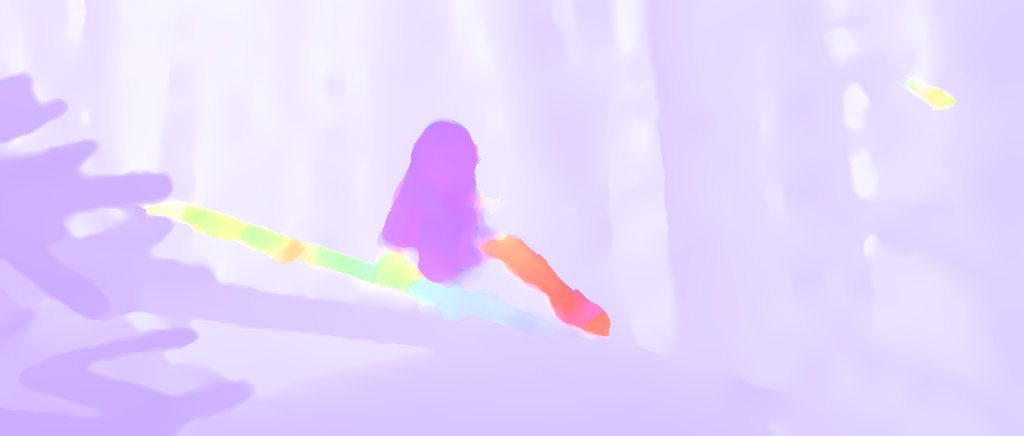}%
		\includegraphics[width=\resultwidth]{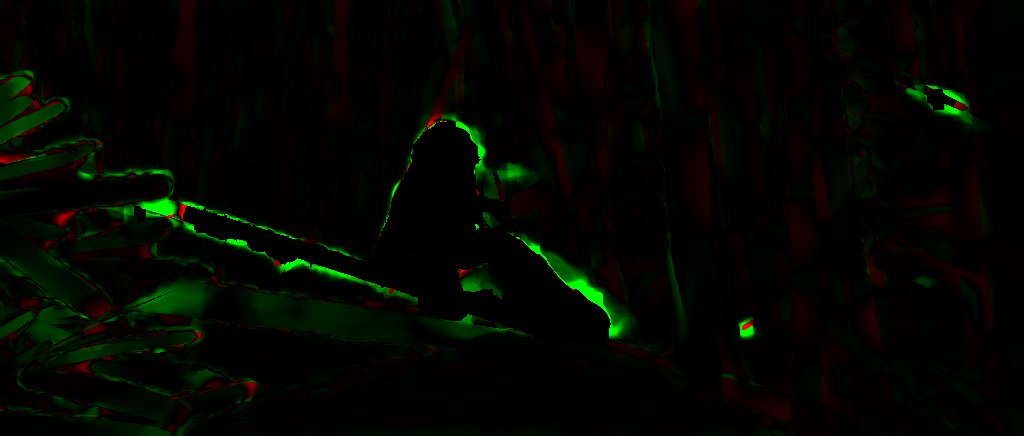}%
	}%
\centerline{
		\includegraphics[width=\resultwidth]{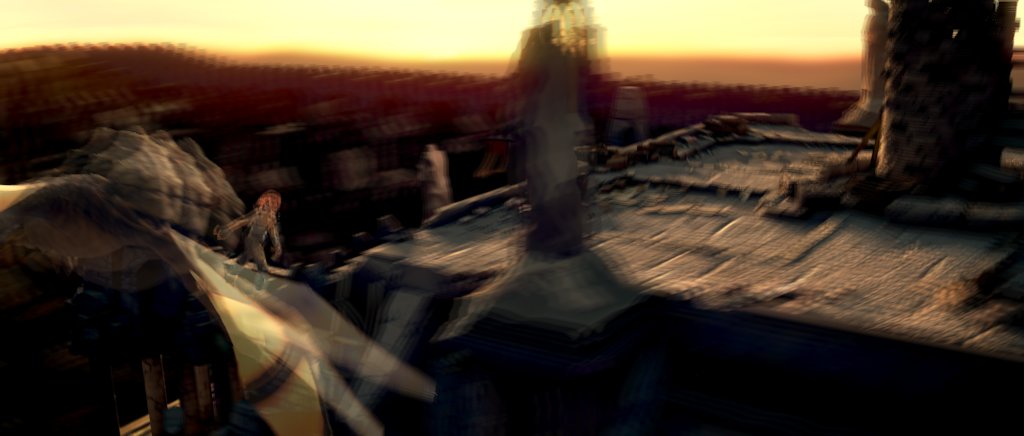}%
		\includegraphics[width=\resultwidth]{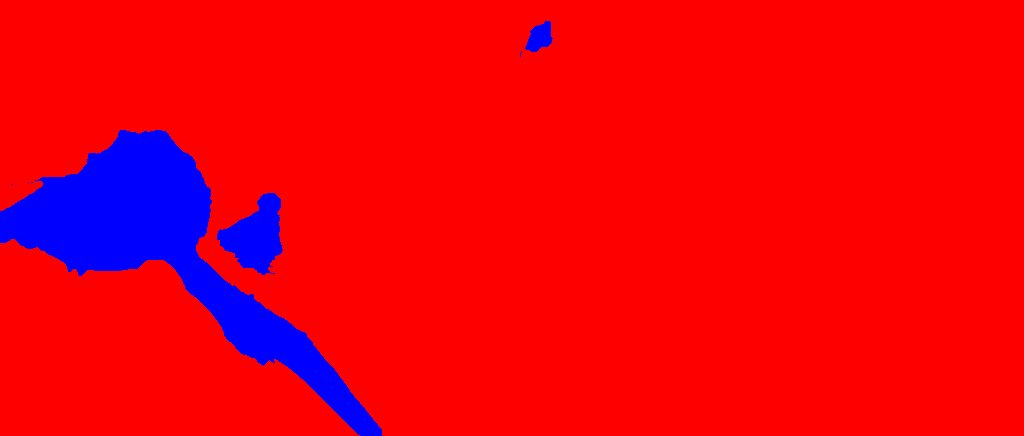}%
		\includegraphics[width=\resultwidth]{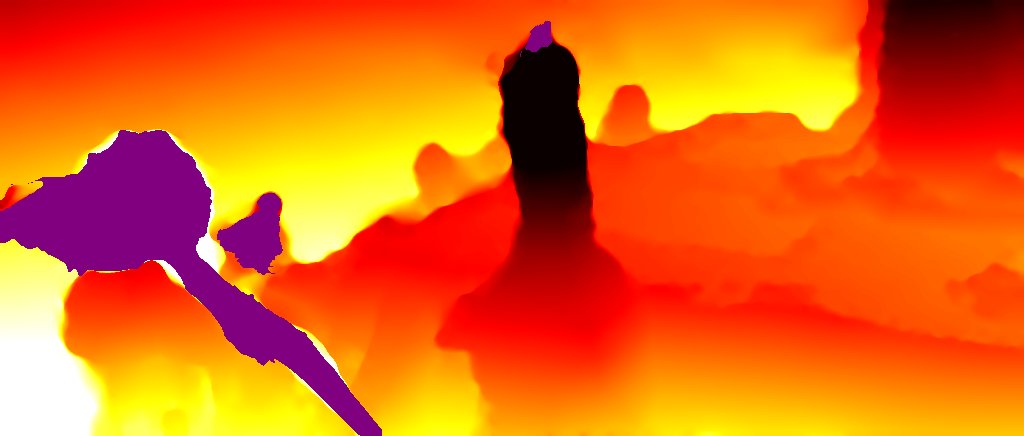}%
		\includegraphics[width=\resultwidth]{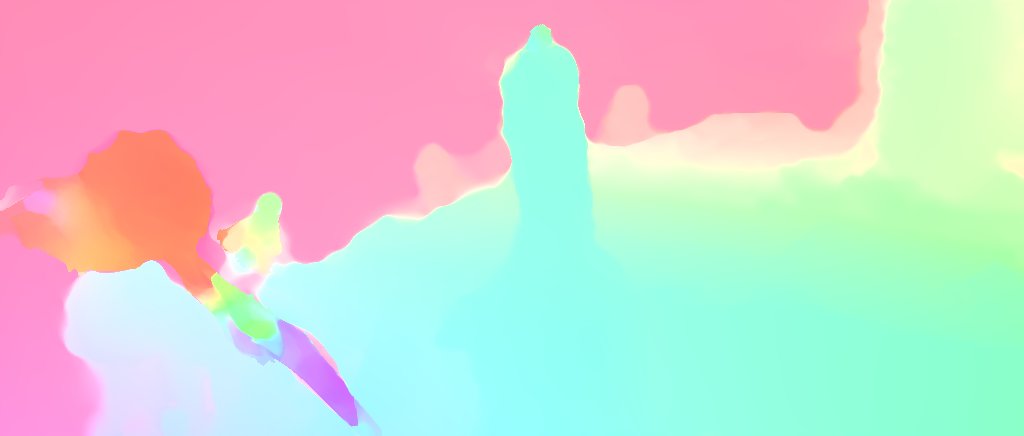}%
		\includegraphics[width=\resultwidth]{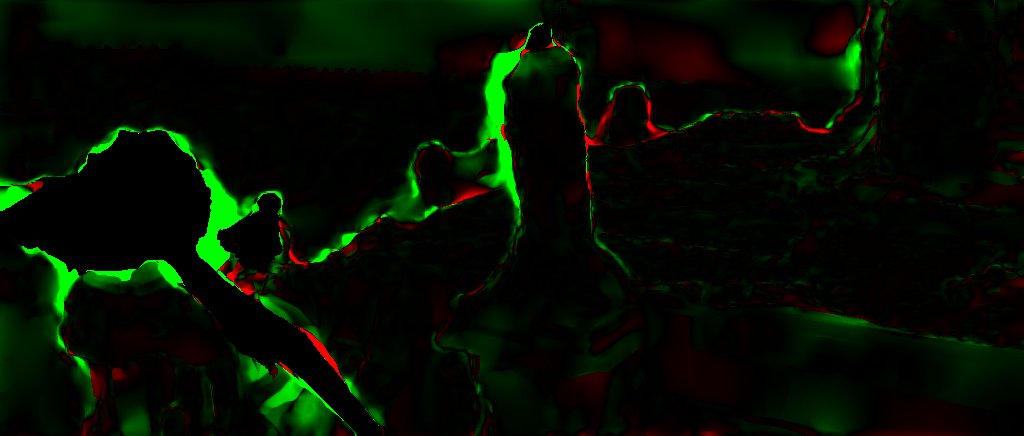}%
	}%
\centerline{
		\includegraphics[width=\resultwidth]{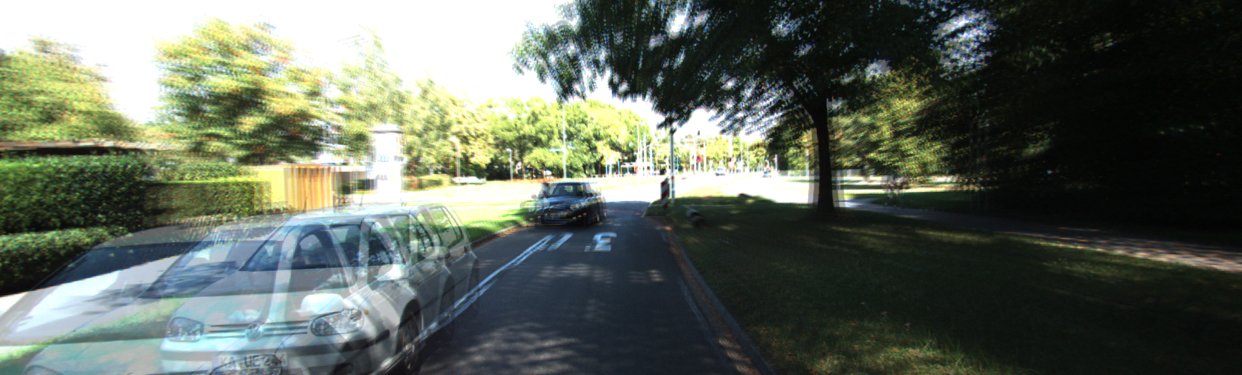}%
		\includegraphics[width=\resultwidth]{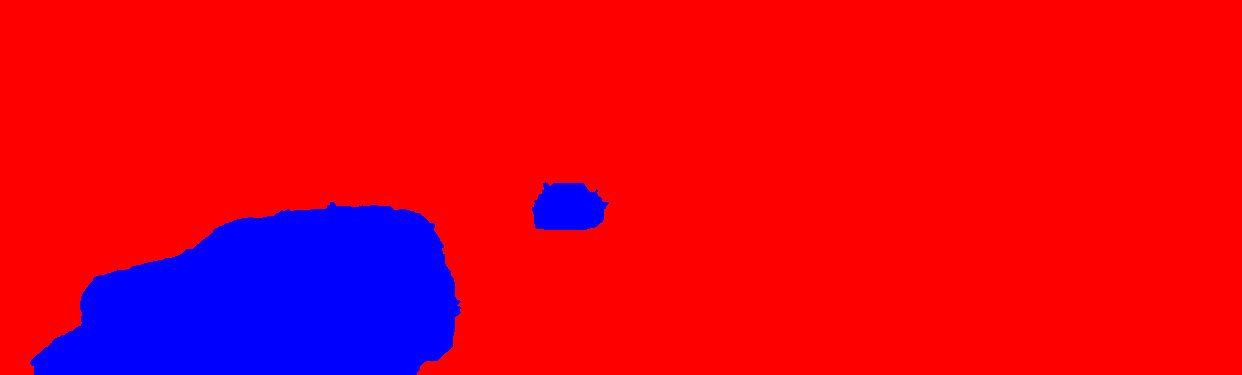}%
		\includegraphics[width=\resultwidth]{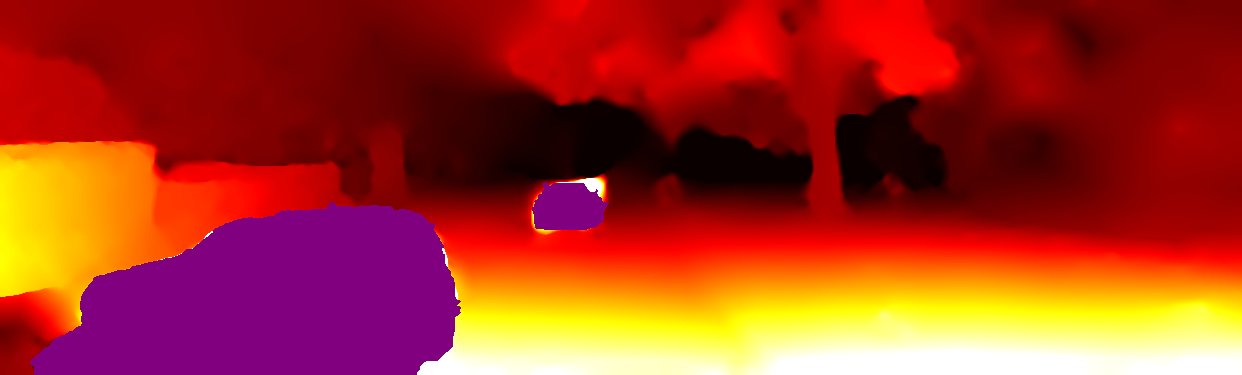}%
		\includegraphics[width=\resultwidth]{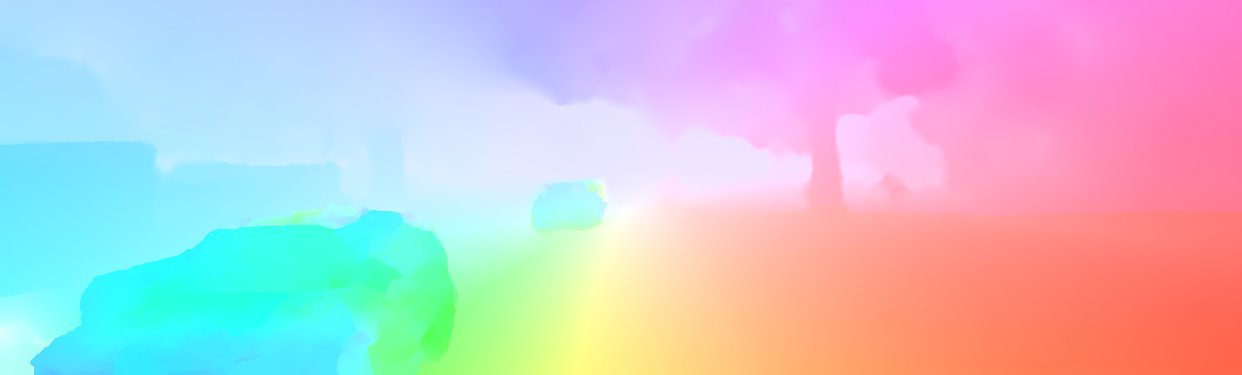}%
		\includegraphics[width=\resultwidth]{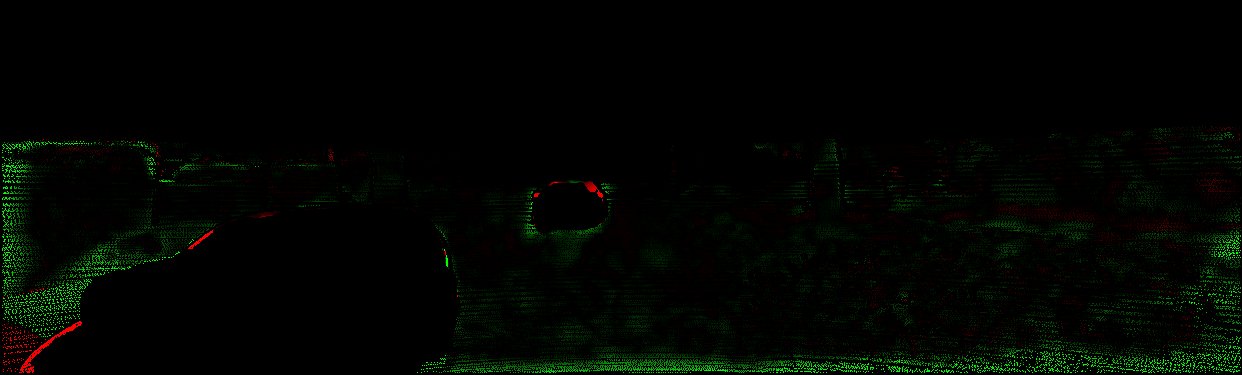}%
	}%
\centerline{
		\includegraphics[width=\resultwidth]{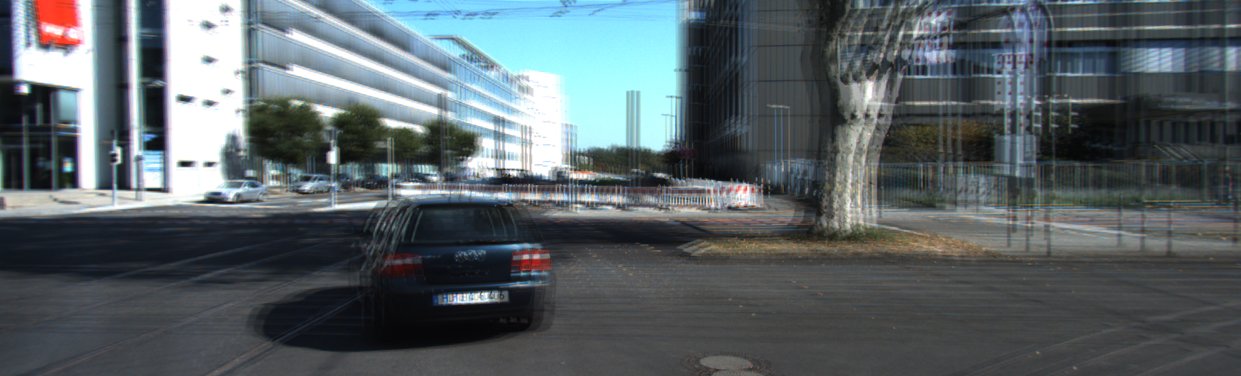}%
		\includegraphics[width=\resultwidth]{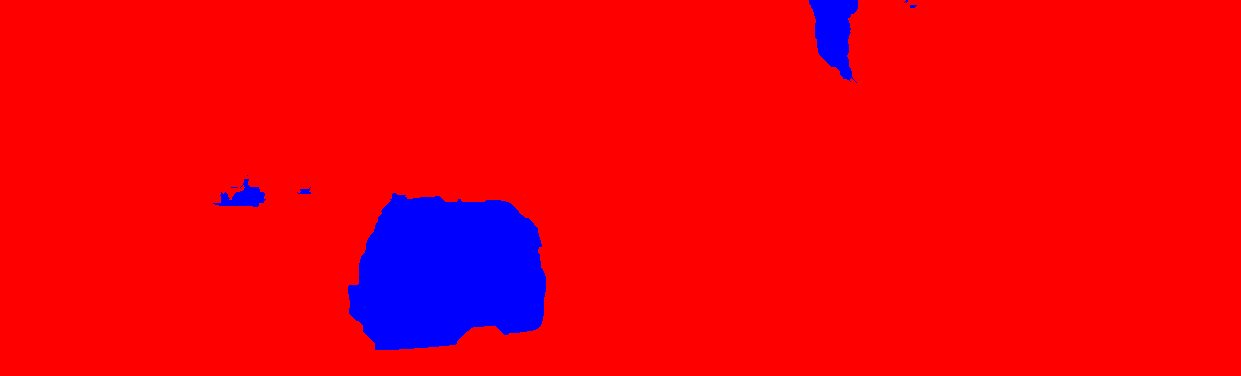}%
		\includegraphics[width=\resultwidth]{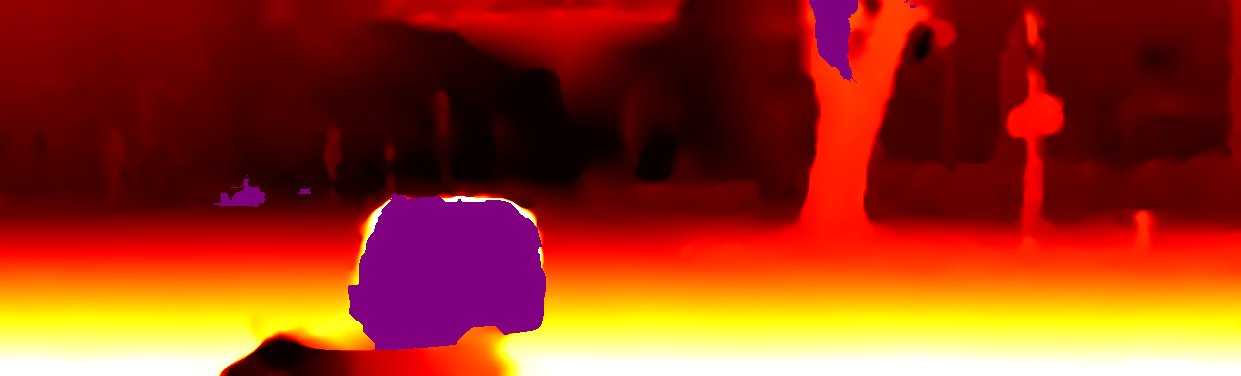}%
		\includegraphics[width=\resultwidth]{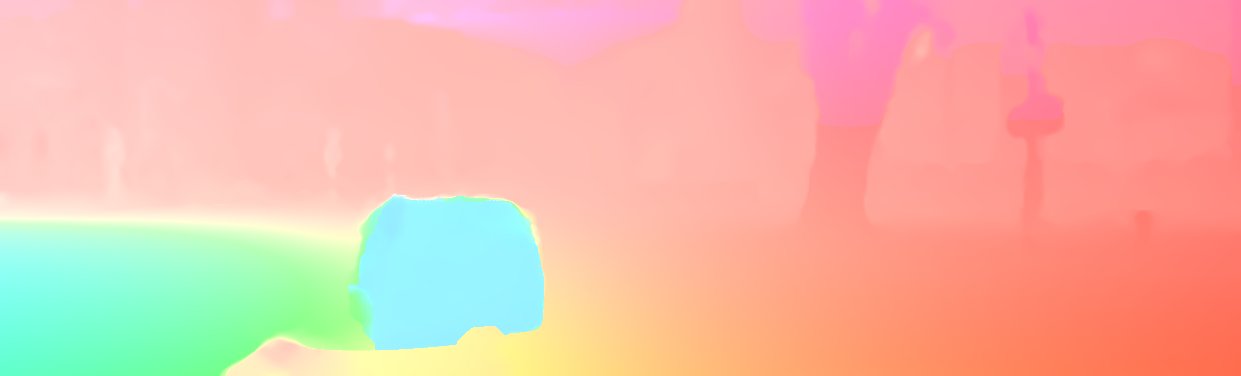}%
		\includegraphics[width=\resultwidth]{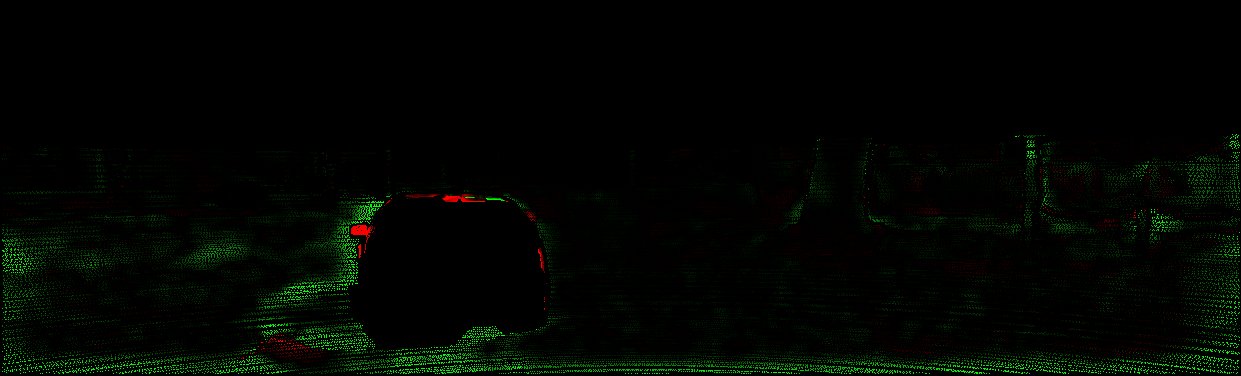}%
	}%
\centerline{
		\includegraphics[width=\resultwidth]{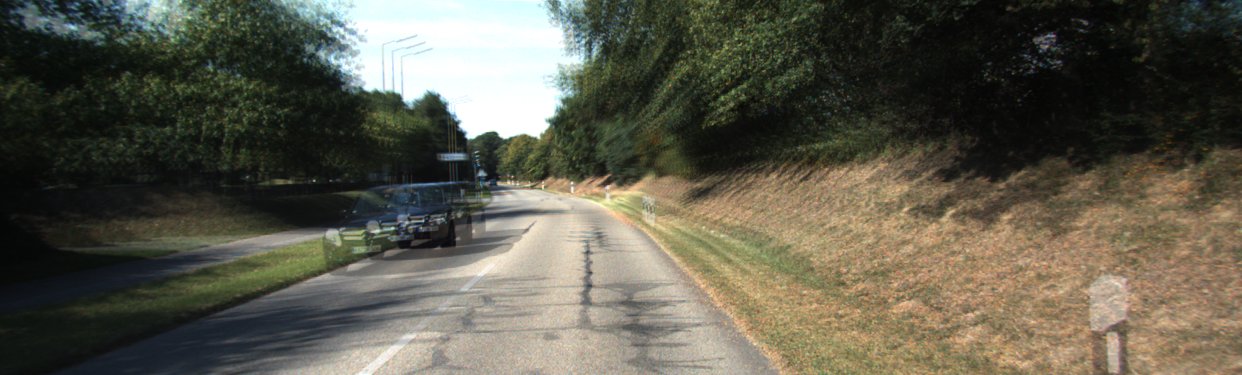}%
		\includegraphics[width=\resultwidth]{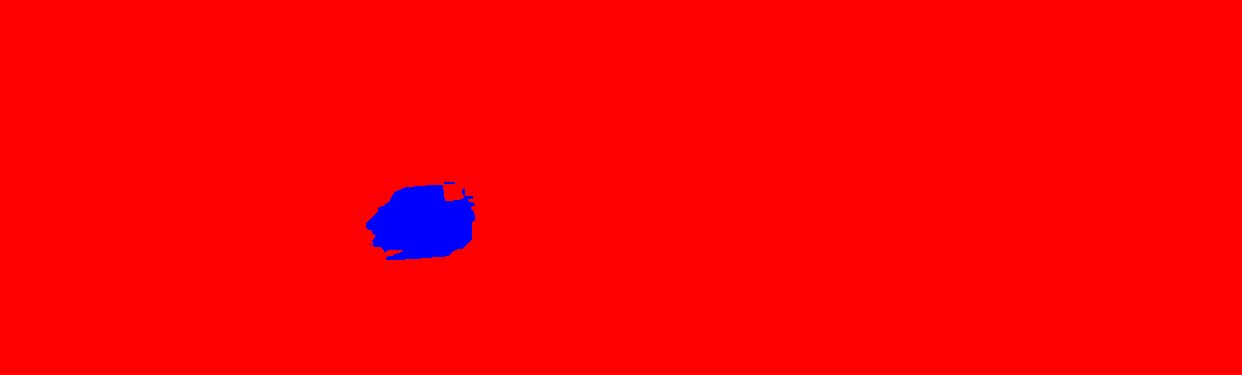}%
		\includegraphics[width=\resultwidth]{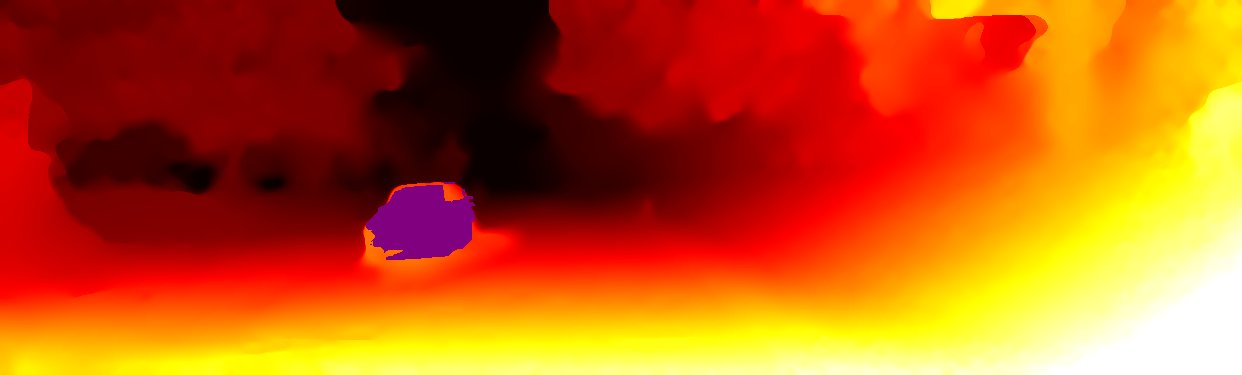}%
		\includegraphics[width=\resultwidth]{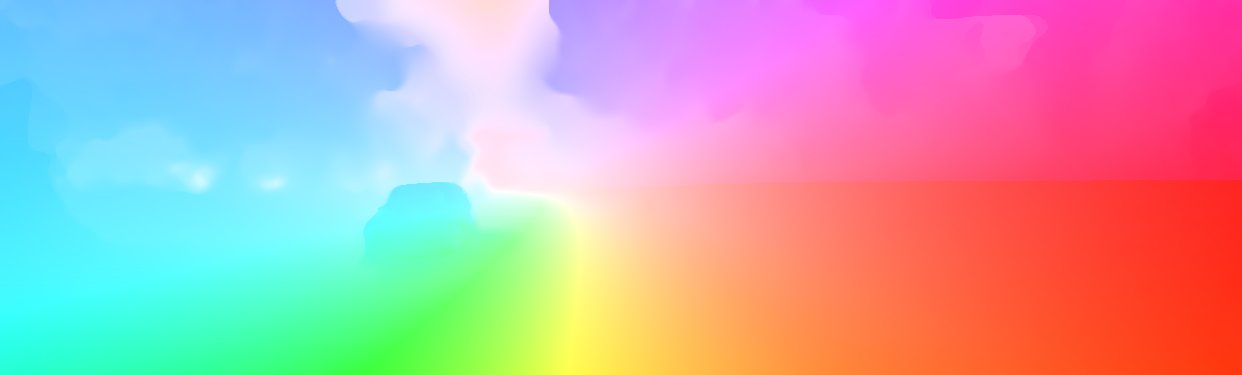}%
		\includegraphics[width=\resultwidth]{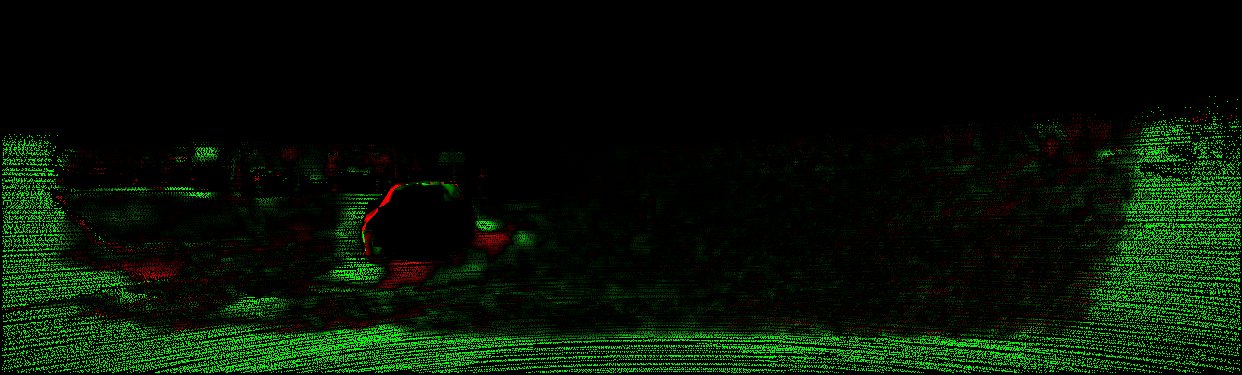}%
	}		
	\caption{Results on MPI-Sintel and KITTI. From left to right: Overlaid input images, rigidity estimation, estimated structure (moving regions are masked in purple), estimated optical flow, comparison to initial flow (green areas denote improvements).} 
	\label{fig:results}
\end{figure*}

\begin{table}
\begin{center}
\setlength{\tabcolsep}{3pt}
\begin{tabular}{lcccccc}
\toprule 
     & \multicolumn{4}{c}{Sintel} & \multicolumn{2}{c}{KITTI 2015} \\
     \cmidrule(lr){2-5} \cmidrule(lr){6-7}
     & \multicolumn{2}{c}{Clean} & \multicolumn{2}{c}{Final} & \\
     \cmidrule(lr){2-3} \cmidrule(lr){4-5}
     & Train & Test  & Train & Test & Train & Test \\
    \cmidrule(lr){2-3} \cmidrule(lr){4-5} \cmidrule(lr){6-7}
     DF~\cite{Menze2015GCPR} & 1.96 & 3.57 & 3.80 & 6.08 & 23.09\% & 21.57\%  \\
     FF+~\cite{Bailer:2015:FlowFields} & - & 3.10 & - & 5.71 & - & - \\
     SDF~\cite{Bai:2016:SemanticDeepFlow} & - & - & - & - &  {\bf 12.14\%} & {\bf 11.01\%} \\
     MR-Flow & {\bf 1.83} & {\bf 2.53} & {\bf 3.59} & {\bf 5.38} & 14.09\% & 12.19\% \\
    \bottomrule
\end{tabular}
\caption{Errors on Sintel (EPE) and KITTI (\%incorrect).}
\label{tab:errors_sintel}
\end{center}
\vspace{-0.3in}
\end{table}

\noindent
To quantify our method, we evaluate on the MPI-Sintel and KITTI-2015 flow benchmarks.
The parameters are chosen to minimize errors on the training sets, and are set to
$\lbrace \sigma_d, \sigma_s, \lambda_{r,c}, \lambda_{r,p}, \lambda_c, \lambda_{1st}, \lambda_{2nd} \rbrace = \lbrace 0.75, 2.5, 0.1, 1.1, 0, 0.1, 5e3 \rbrace$ for Sintel and $\lbrace 1.0, 0.25,$ $0.5, 1.1, 0.01, 1, 5e4 \rbrace$ for KITTI.
Table~\ref{tab:errors_sintel} shows the errors for our method,
our initialization (DF), and for top
performing methods on MPI-Sintel (FF+)~\cite{Bailer:2015:FlowFields} and
KITTI-2015 (SDF)~\cite{Bai:2016:SemanticDeepFlow}. 
Both evaluate only on one dataset; in contrast, our method achieves high accuracy on both datasets.
Figure~\ref{fig:results} visualizes results;
for more results see \cite{MRFlow:Website}.

%

On \textbf{MPI-Sintel}, our method currently outperforms all published works. 
In particular, the structure estimation gives flow in occluded
regions, producing the {\it lowest errors in the unmatched regions}
of any published or unpublished work.
On a 2.2 GHz i7 CPU, our method takes on average 2 minutes per triplet of frames without the initial flow computation, 74s for the initialization and rigidity estimation, and 46s for the optimization.



In \textbf{KITTI-2015}
the scenes are simpler and contain only automotive situations; however, the images suffer from artifacts such as noise and overexposures.
Among published monocular methods, MR-Flow is second
after~\cite{Bai:2016:SemanticDeepFlow}, which is designed for
automotive scenarios and not tested on Sintel.

\section{Conclusion}
\noindent
We have demonstrated an optical flow method that segments the scene
and  improves accuracy by exploiting rigid
scene structure.
We combine semantic and motion information to detect independently moving regions, and use an existing flow method to compute the motion of these regions.
In rigid regions of the scene, the flow is directly constrained by the 3D structure of the world.
This allows us to implicitly regularize the flow by constraining the underlying structure to a locally planar model.
Furthermore, since the structure is temporally coherent, we combine information from multiple frames.
We argue that this uses the right constraints in the right place and
produces accurate flow in challenging situations and competitive
results on Sintel and KITTI.

This opens several directions for future work.
First, the rigidity estimation could be improved using better
inference algorithms and training data.
Jointly refining the foreground flow with the rigid flow estimation could improve
performance.
Our method could also use longer sequences, and enforce temporal
consistency of the rigidity maps.

\noindent
{\small \textbf{Acknowledgements.} JW and LS were supported by the Max Planck ETH Center for Learning Systems.}

\pagebreak

{\small
\bibliographystyle{ieee}
\bibliography{pppflow}
}

\pagebreak
\appendix
\section{Supplemental Material}
\subsection{Semantic rigidity estimation: CNN architecture details and training procedure}
\label{sec:cnndetails}
\begin{figure*}[t]
 \centerline{
 	\includegraphics[width=0.195\textwidth]{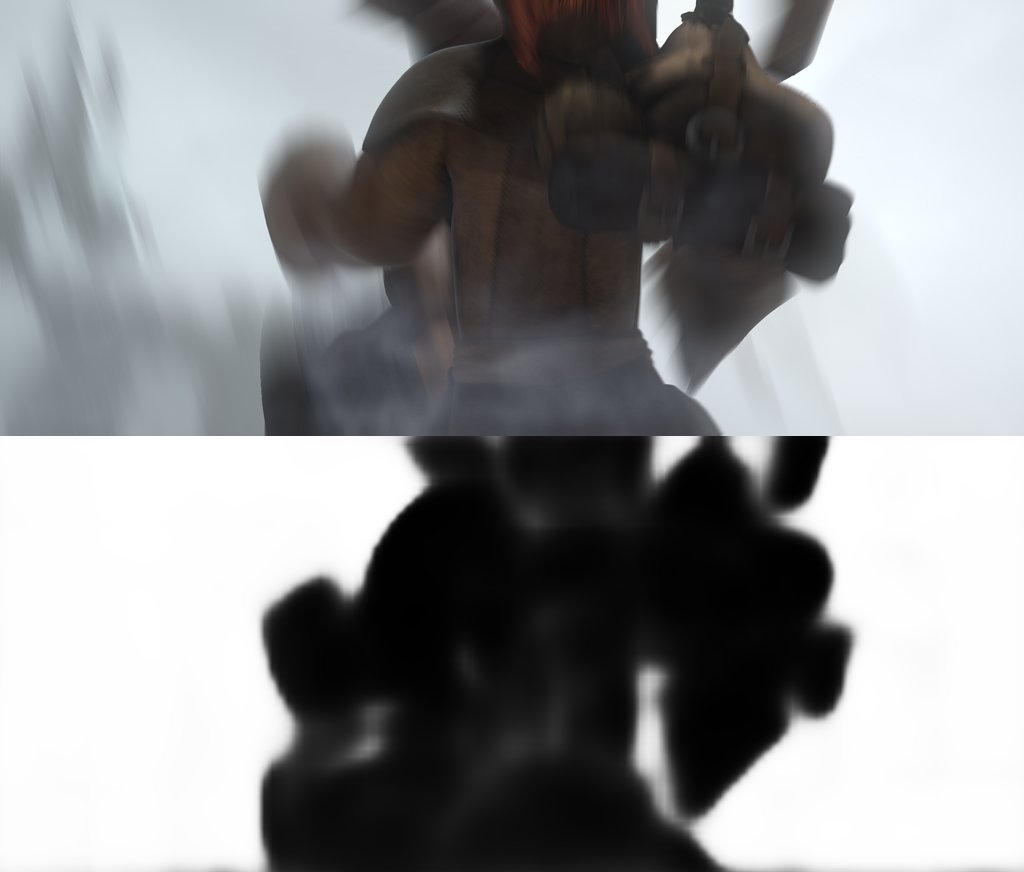}
 	\includegraphics[width=0.195\textwidth]{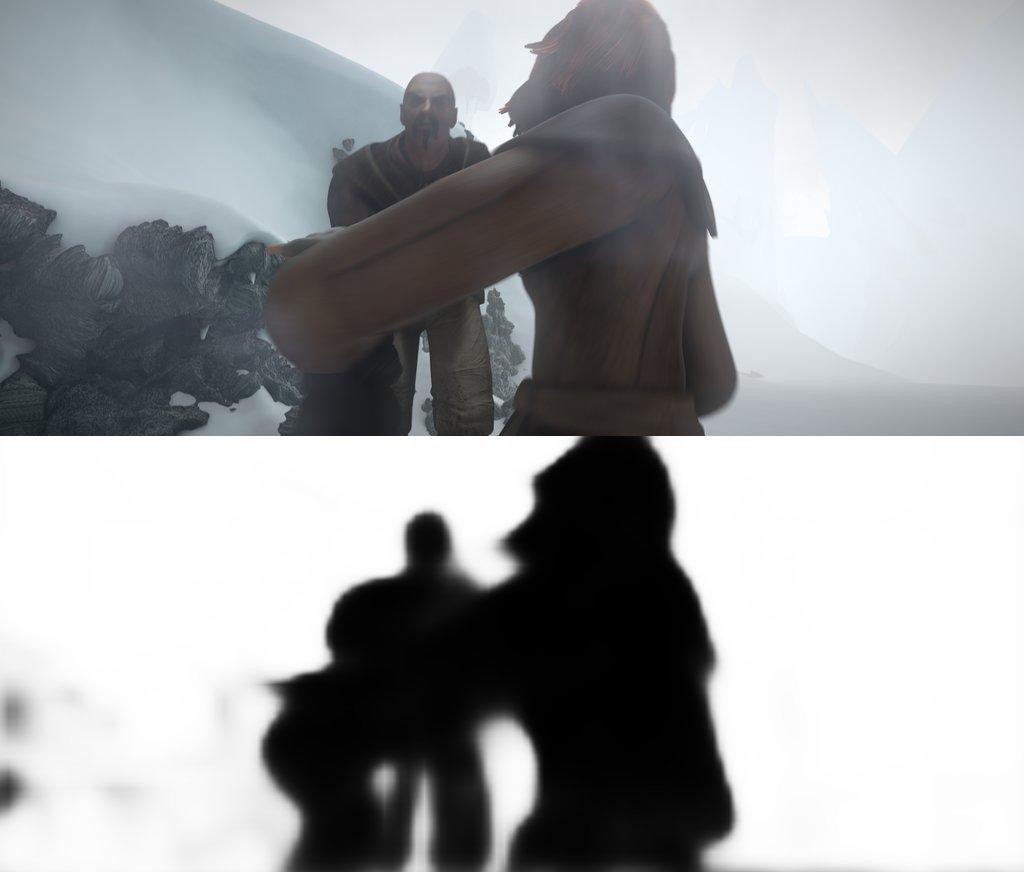}		
 	\includegraphics[width=0.195\textwidth]{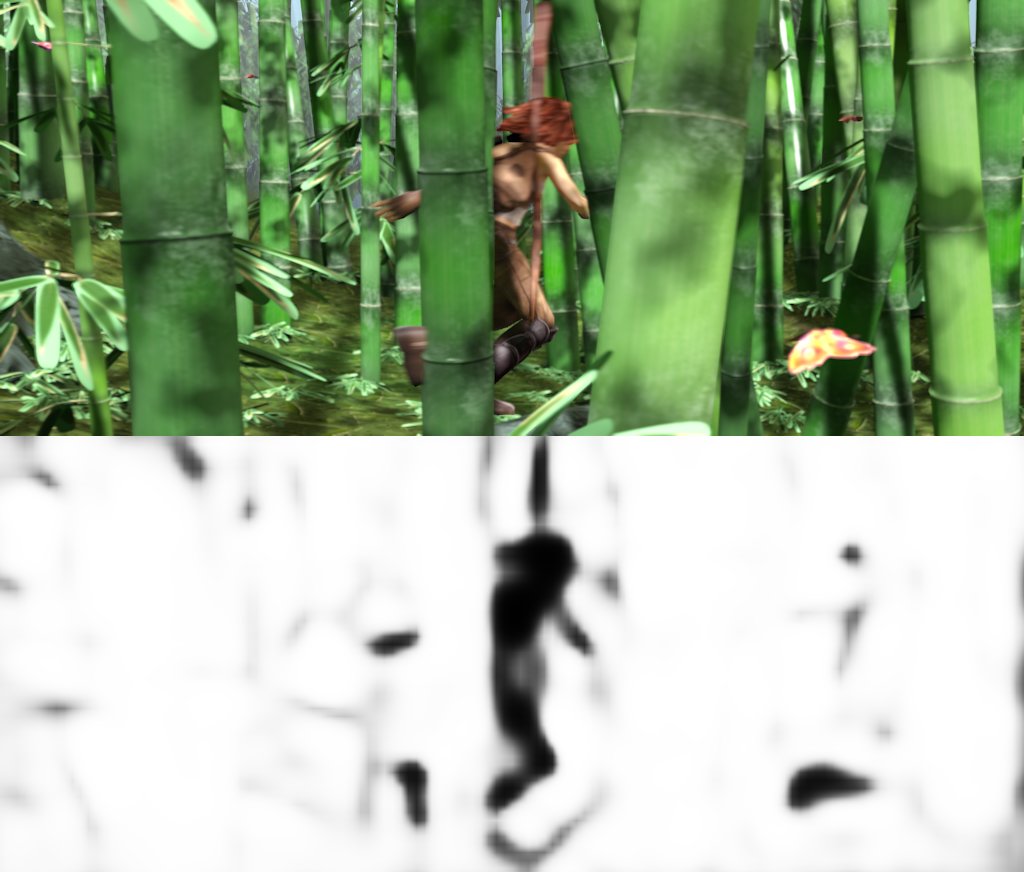}		
 	\includegraphics[width=0.195\textwidth]{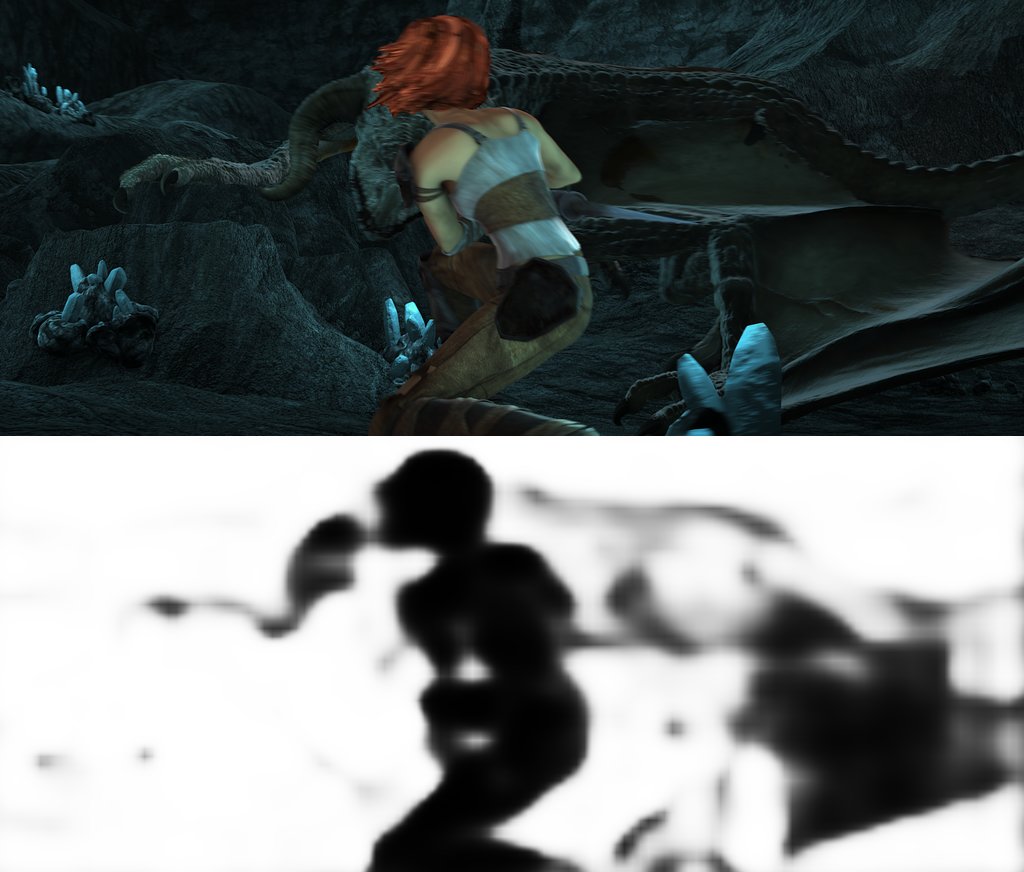}		
 	\includegraphics[width=0.195\textwidth]{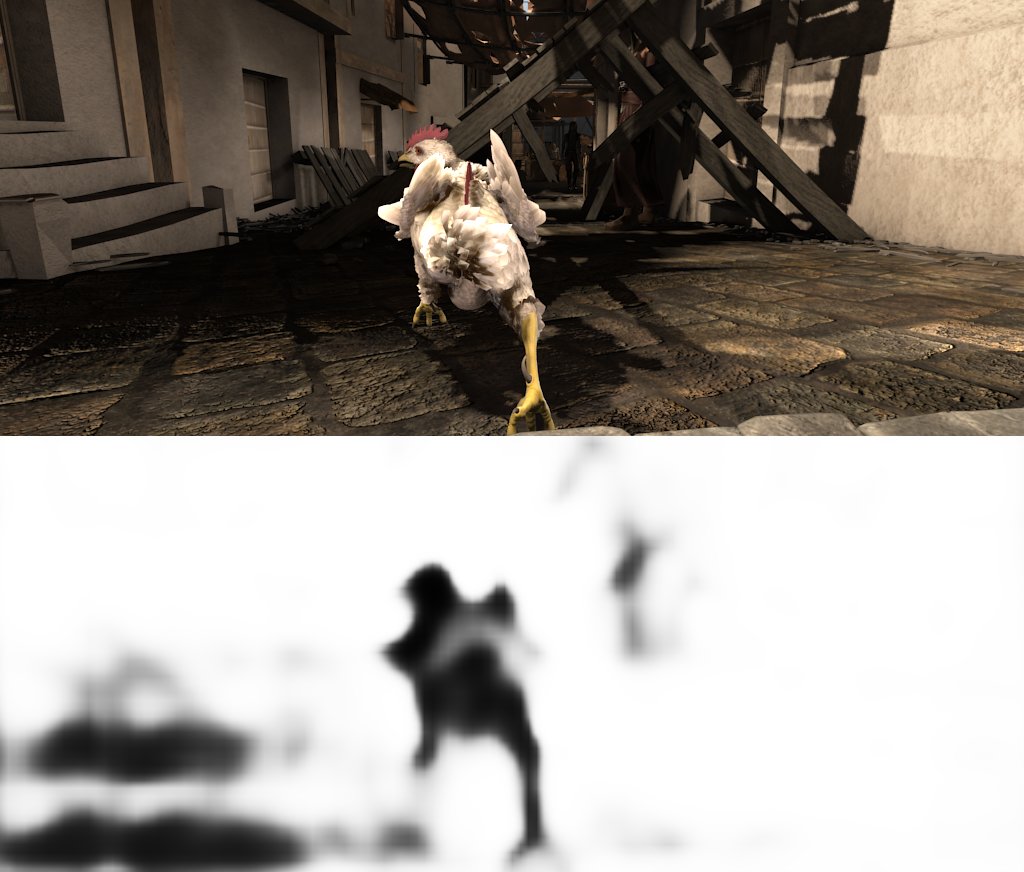}		
 }
 \vspace{-0.05in}
 \caption{Results of rigidity estimation on the MPI-Sintel test set. Top: Original frame; Bottom: probability map of rigidity (white is likely rigid, black likely moving independently). }
 \label{fig:rigidity_sintel}
 \end{figure*}

\begin{figure*}[t]
 \centerline{
 	\includegraphics[width=0.245\textwidth]{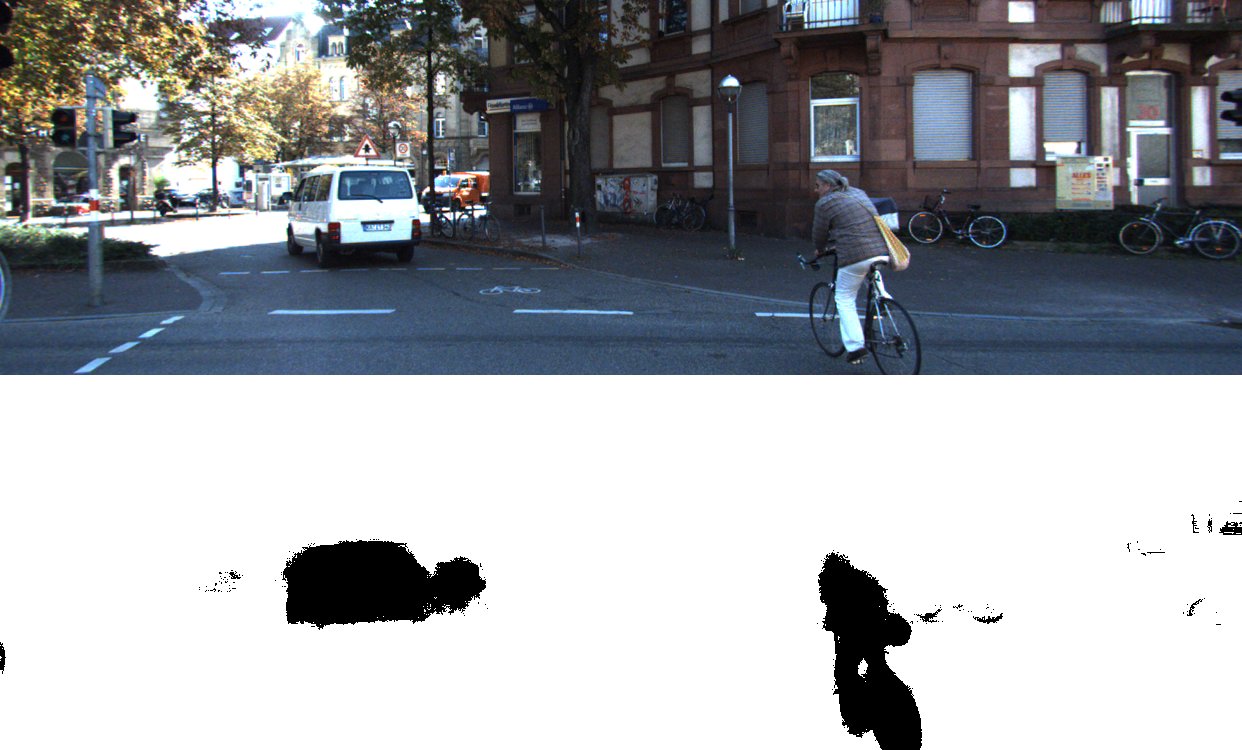}
 	\includegraphics[width=0.245\textwidth]{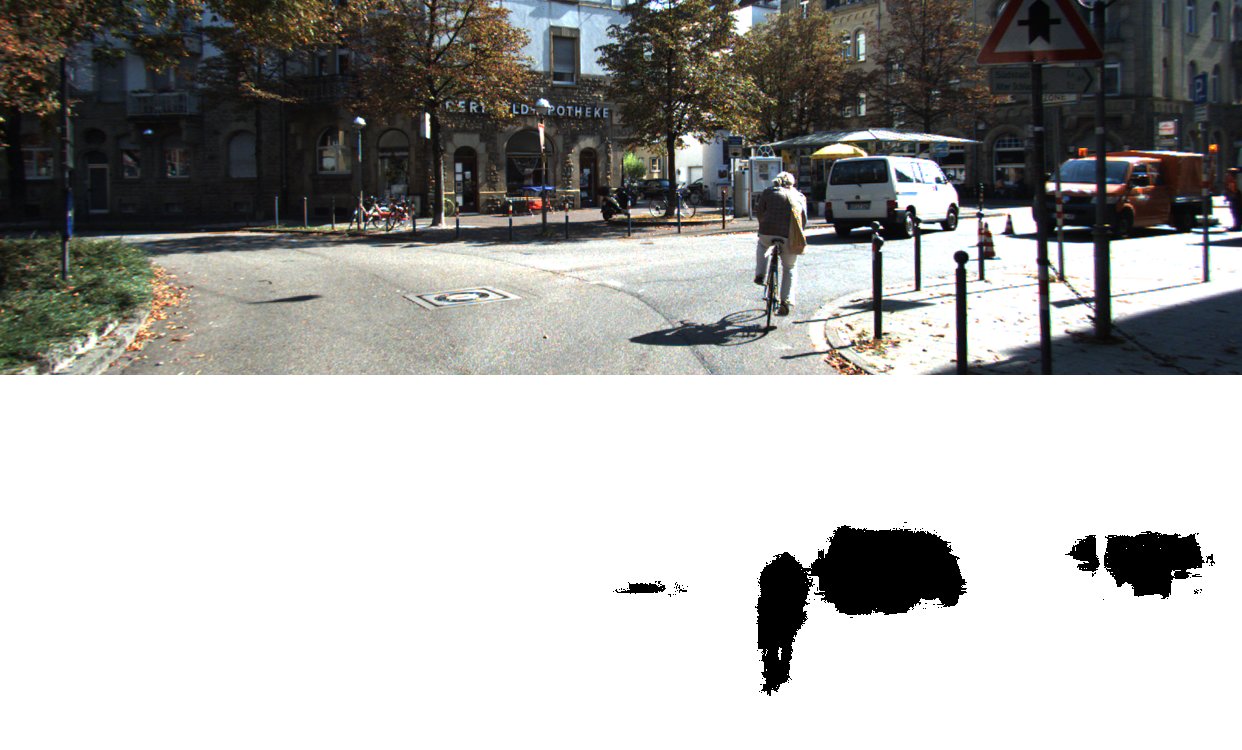}		
 	\includegraphics[width=0.245\textwidth]{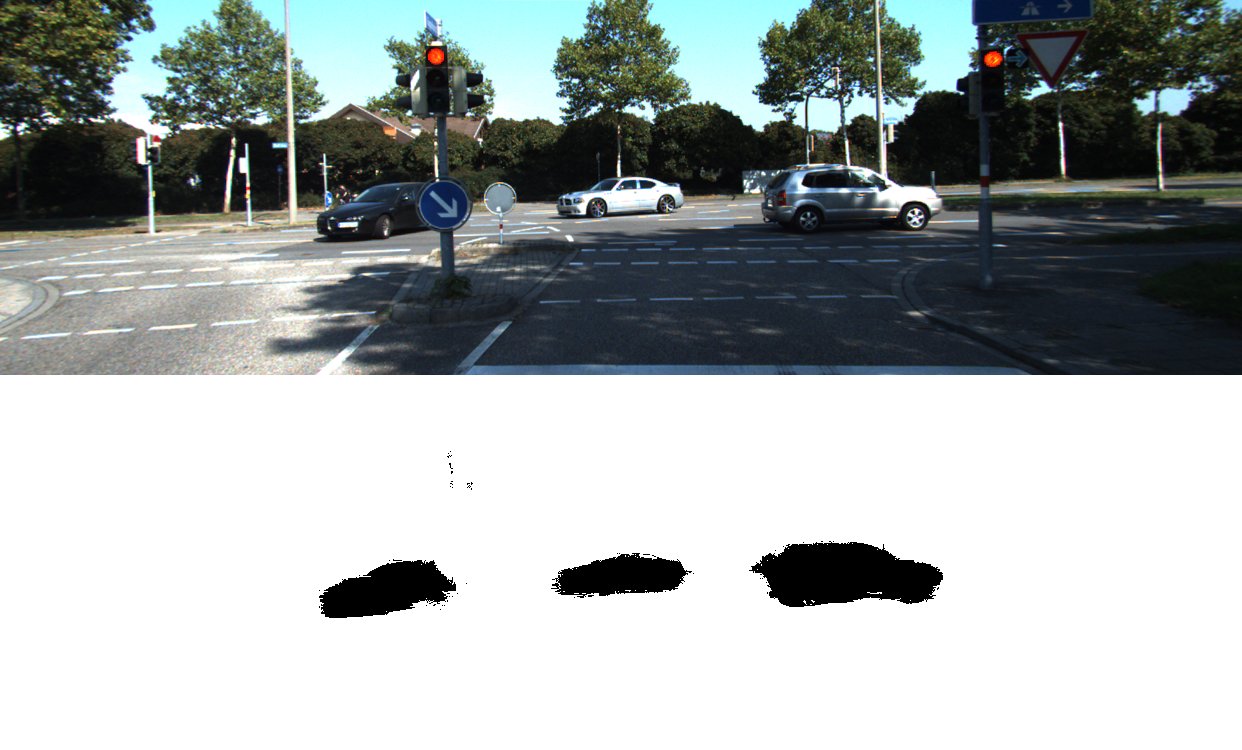}		
 	\includegraphics[width=0.245\textwidth]{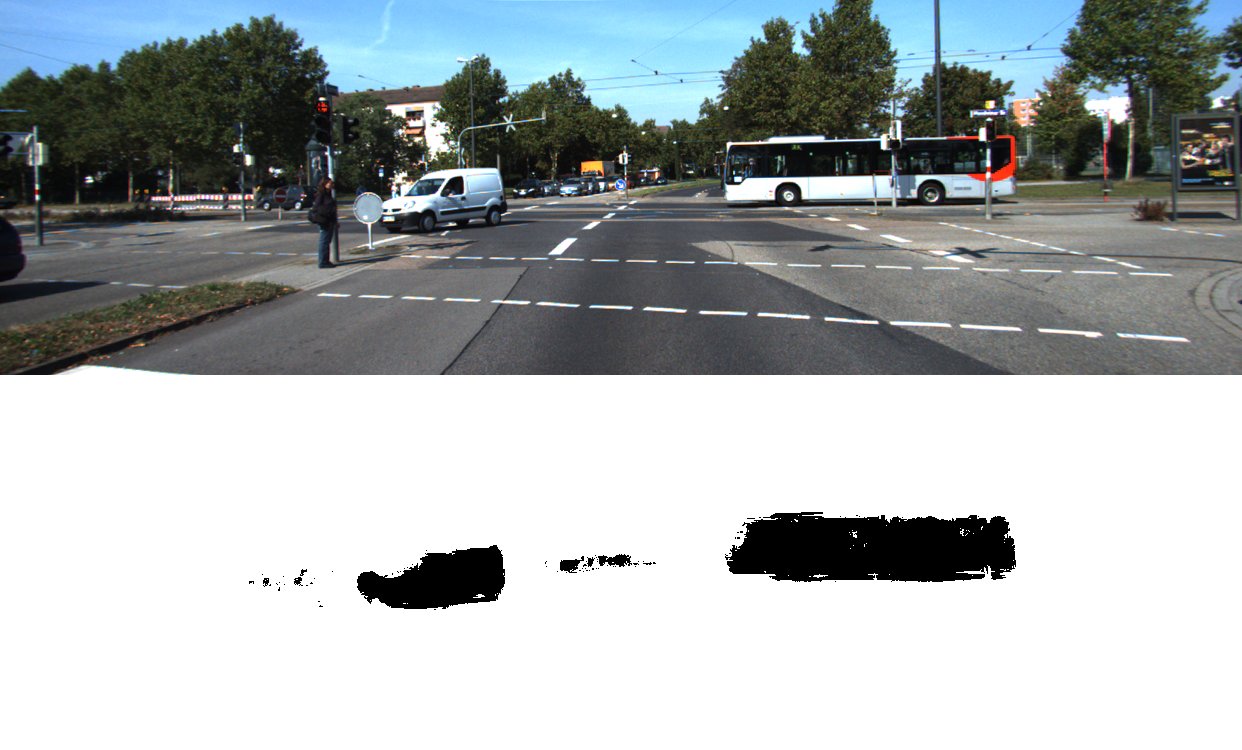}	
 }
 \vspace{-0.1in}
 \caption{Results of rigidity estimation on KITTI 2015. Top: Original frame; Bottom: probability map of rigidity. }
 \label{fig:rigidity_kitti}
 \end{figure*}

As described in our paper, each of the two datasets provide different data, and thus we used slightly different procedures to estimate rigidity on each of the datasets. Here we describe the details.
Additionally, Fig.~\ref{fig:rigidity_sintel} and Fig.~\ref{fig:rigidity_kitti} show some more examples of the outputs of the networks on images that were not seen during test time.

{\bf KITTI 2015:} We followed the same procedure as previous work on semantic segmentation on KITTI~\cite{Sevilla:SOF}, since it was shown to be successful. We used the DeepLab architecture~\cite{ChenPKMY14} modifying the output layer to classify 22 classes: aeroplane, bicycle, bird, boat, building, bus, car, cat, cow, dog, floor, grass, horse, motorbike, road, sheep, sidewalk, sky, train, water, person and background, where background includes anything that is not one of the other 21 classes. We initialized the weights with the VGG model trained on Pascal VOC and then fine-tuned it on our categories using a fixed momentum of 0.9, weight decay of 0.0005 and learning rate of 0.0001 for the first 100K iterations, reduced by 0.1 after every 50K steps, during 200K iterations. We used a dense CRF~\cite{Krahenbuhl:2011:DenseCRF} on top, where the unaries are the CNN output at each pixel and the pairwise potentials are position and a bilateral kernel with both position and RGB values. The inference in the dense CRF model is performed using 10 steps of mean-field approximate inference. At test time we obtain a probability over the classes. We estimate rigidity by choosing the class with the highest probability, and classifying the pixel as rigid or non-rigid based on whether an object in the class is capable of moving independently (for example, car) or not (for example, building). The accuracy of rigidity classification on the training set is 96.09\%, where rigid parts are correctly classified 96.93\% of the time, and independent moving parts are correctly classified 91.51\% of the time. 

{\bf MPI-Sintel:} In this dataset there is no previous work on estimating rigidity. Thus, we used one of the latest released versions of the DeepLab architecture, called DeepLab-Coco-LargeFov, which is pretrained including extra annotations from the MS-COCO dataset\footnote{Further details can be found on their webpage~\url{http://ccvl.stat.ucla.edu/software/deeplab/deeplab-coco-largefov/}}. We modified the output layer to classify pixels as rigid or nonrigid. We fine-tuned all layers using the same parameters as before for 1.4K iterations. This small number of iterations was selected to avoid overfitting. At test time, we obtain a probability of rigidity, and we compute the final estimate of rigidity by thresholding at 0.5. The accuracy of the estimation on a validation set is 94.2\%.

%
%
\subsection{Derivation of direction-based rigidity likelihood}
\label{sec:rigiditydetails}

While the accuracy of the CNN-based rigidity estimation is surprisingly high, there are still occasions when it fails.
This may happen for example when there is strong motion blur in the rigid regions, causing these to be classified as independently moving, which is a false positive. 
On the other hand, a noisy or over-saturated appearance of objects can cause the semantic segmentation to fail and result in a false negative.
Therefore, we combine the CNN-based rigidity estimation with a motion-based estimation, described in this section.

\begin{figure}
	\centering
	\includegraphics[width=0.45\textwidth]{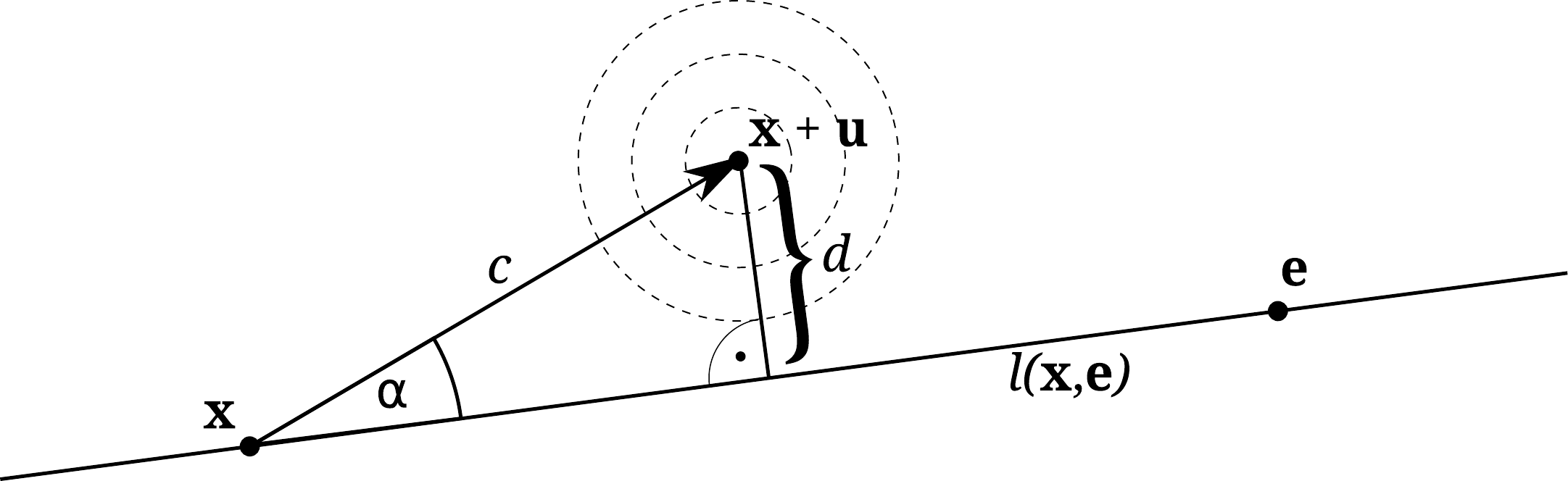}
	\caption{Illustration}
	\label{fig:illustration}
\end{figure}

For a given point $\mathbf{x}$, our model assumes the measured corresponding point $\mathbf{x}' = \mathbf{x} + \mathbf{u}$ to have a Gaussian error distribution around the true correspondence with covariance matrix $\Sigma = \sigma^2 \mathbf{I}$.
For a given rigid point (denoted by the conditioning on $r=1$) and a given focus of expansion $\mathbf{e}$, the probability of the true correspondence pointing towards $\mathbf{e}$ is then given as
\begin{equation}
p \left( \mathbf{x'} \vert r=1, \mathbf{e} \right)
=
\int_{y \in l(\mathbf{x},\mathbf{e} ) }
\frac{1}{2 \pi \sigma_m^2} \exp \left( - \frac{1}{2} \mathbf{y^T} \Sigma^{-1} \mathbf{y} \right) d\mathbf{y},
\label{eq:lineintegral}
\end{equation}
where $l(\mathbf{x},\mathbf{e} )$ denotes the line that goes through $\mathbf{x}$ and $\mathbf{e}$.
Figure~\ref{fig:illustration} shows an illustration.

Since the error distribution is Gaussian, every marginal is also Gaussian.
Therefore, the line integral in \eqref{eq:lineintegral} is given as
\begin{equation}
p \left( \mathbf{x'} \vert r=1, \mathbf{e} \right)
=
\frac{1}{\sqrt{2 \pi \sigma_m^2}}
\exp \left( - \frac{d^2}{2 \sigma_m^2} \right),
\label{eq:prob_xq}
\end{equation}
where $d$ denotes the distance of $\mathbf{x'}$ to $l(\mathbf{x},\mathbf{e} )$, as shown in Figure~\ref{fig:illustration}.

In the following, it will be convenient to express the correspondence point $\mathbf{x'}$ in \eqref{eq:prob_xq} in terms of the angle $\alpha$ and the displacement magnitude $c = \Vert \mathbf{u} \Vert$.
Then,
\begin{equation}
p \left( \alpha, c \vert r=1, \mathbf{e} \right)
=
\frac{1}{\sqrt{2 \pi \sigma_m^2}}
\exp \left( - \frac{c^2 \sin^2(\alpha)}{2 \sigma_m^2} \right) .
\label{eq:prob_calpha}
\end{equation}
We can thus drop $\mathbf{e}$ from the equation.

The likelihood of a point being rigid given the measurements $\alpha, c$ is now
\begin{align}
& p(r=1 \vert \alpha, c) \nonumber \\
&=
\frac{ p(\alpha \vert r=1, c) p(r=1)}
{p(\alpha \vert r=0, c) p(r=0) + p(\alpha \vert r=1, c) p(r=1) } \nonumber \\
&=
\frac{ \frac{1}{Z} p(\alpha, c \vert r=1) p(r=1)}
{p(\alpha \vert r=0, c) p(r=0) + \frac{1}{Z} p(\alpha, c \vert r=1) p(r=1) } .
\end{align}
Abbreviating the prior for rigidity $p(r=1)$ as $p_1$ and setting $p(\alpha \vert r=0, c) = \frac{1}{2\pi}$ (the motion of independently moving regions is supposed to be uniformly distributed), we get
\begin{equation}
p(r=1 \vert \alpha, c) = 
\frac{ p_1 p(\alpha, c \vert r=1)}
{\frac{Z (1-p_1)}{2\pi} + p_1 p(\alpha, c \vert r=1)}
\end{equation}
and for uninformative priors $p_1 = 0.5$
\begin{equation}
p(r=1 \vert \alpha, c) = 
\frac{ p(\alpha, c \vert r=1)}
{\frac{Z}{2\pi} + p(\alpha, c \vert r=1)} .
\label{eq:p_rigid}
\end{equation}

What remains is to compute $Z$.
Using Eq.~\eqref{eq:prob_calpha}, it can be computed as
\begin{align}
Z &= p(c \vert r=1) = \int_{0}^{2\pi} p(\alpha, c \vert r=1 ) \mathrm{d}\alpha \nonumber \\
&= \frac{1}{\sqrt{2 \pi \sigma_m^2}} \int_{0}^{2\pi} \exp \left( - \frac{c^2 \sin^2(\alpha)}{2 \sigma_m^2} \right) \mathrm{d}\alpha \nonumber \\
& \left[ \Scale[0.75]{t = \frac{c^2}{4\sigma_m^2}, \sin^2(x) = \frac{1}{2}(1-\cos(2x))} \right] \nonumber \\
&= \frac{1}{\sqrt{2 \pi \sigma_m^2}} \int_{0}^{2\pi} \exp \left( - t (1-\cos(2\alpha)) \right) \mathrm{d}\alpha \nonumber \\
&= \frac{1}{\sqrt{2 \pi \sigma_m^2}} \exp\left( -t \right) \int_{0}^{2\pi} \exp \left( t \cos(2\alpha) \right) \mathrm{d}\alpha\nonumber \\
& \left[ \Scale[0.75]{\beta = 2\alpha} \right] \nonumber \\
&= \frac{1}{\sqrt{2 \pi \sigma_m^2}} \exp\left( -t \right) 
	\frac{1}{2} \int_{0}^{4\pi} \exp \left( t \cos(\beta) \right) \mathrm{d}\beta \nonumber \\
&= \frac{1}{\sqrt{2 \pi \sigma_m^2}} \exp\left( -t \right) 
	\int_{0}^{2\pi} \exp \left( t \cos(\beta) \right) \mathrm{d}\beta \nonumber \\
&= \frac{1}{\sqrt{2 \pi \sigma_m^2}} \exp\left( - t \right) 
	2\pi \mathbb{I}_0 \left( t \right) \nonumber \\
&= \frac{\sqrt{2 \pi}}{\sigma_m} \exp\left( - t \right) 
	\mathbb{I}_0 \left( t \right), 
\label{eq:z}
\end{align}
with $\mathbb{I}_0(x)$ the modified Bessel function of the first kind.
Inserting Eq.~\eqref{eq:z} and Eq.~\eqref{eq:prob_calpha} into Eq.~\ref{eq:p_rigid} yields Eq.~(\textcolor{red}{11}) from the main paper:
\begin{align}
& p\left(\mathbf{x} \text{ is rigid} \right) \nonumber \\
&= 
p(r=1 \vert \alpha, c) \nonumber \\
&= \frac{\exp \left( - 2 t \sin^2 ( \alpha ) \right)}
{\exp(-t) \mathbb{I}_0(t) + \exp \left( -2t \sin^2 \left( \alpha \right) \right)}.
\end{align}
%


\section{Ablation study}
This section provides an additional ablation study that had to be omitted from the paper due to space limitations.
Here, we test how different subcomponents of our algorithm impact the end result\footnote{The results in this section were obtained using a reduced version of the MPI-Sintel training set containing every 4th frame, and only the \texttt{final} pass.}.
To assess the impact in different regions of the frame, we provide the errors both on the full frames and only in the ground truth rigid regions.

For the ablation study, we successively switch on four steps: \textit{occlusion reasoning}, \textit{coplanarity refinement}, \textit{nonlinear initialization of $\hat{b}^-$}, and \textit{spatial priors}.

\begin{itemize}
\item \textbf{Occlusion reasoning} refers to the estimation of the visibility maps $V^+, V^-$ using the forward-backward consistency, as described in Section 4 in the paper.
If this step is switched off, we set $V^- = V^+ = 1$ everywhere, and therefore do not explicitly exclude occluded pixels from subsequent computations.
\item \textbf{Coplanarity refinement} refers to the second part of the initial alignment (Eq.~(5) in the main paper)\footnote{All equation numbers refer to equations in the main paper.}. This step refines the initial homographies $\bar{H}^-,\bar{H}^+$ to ensure that after registration all residual flow vectors meet in the two epipoles $\mathbf{e}^-, \mathbf{e}^+$.
The optimization yields $\hat{H}^-,\hat{H}^+$.
If this step is switched off, we simply set $\hat{H}^+ = \tilde{H}^+, \hat{H}^- = \tilde{H}^-$.
\item \textbf{Nonlinear initialization of $\hat{b}^-$} refers to initializing $\hat{b}^-$ using Eq.(8), \ie choosing $\hat{b}^-$ so that the resulting backward $\hat{A}^-$ structure is as similar as possible under a robust error norm to the initial forward structure $\hat{A}^+$.
Note that, without loss of generality, $\hat{b}^+$ is always chosen so that the MAD of $\hat{A}^+$ is 1.

If this step is switched off, use Eq.~(7) to equate $\hat{A}^+$ and $\hat{A}^-$.
This produces an estimate of $\hat{b}^-$ per pixel, of which we take the median to arrive at the global estimate of $\hat{b}^-$.
\begin{equation}
\hat{b}^- = \median_\mathbf{x} \frac{1}{A^+(\mathbf{x})} \frac{w^-(\mathbf{x})}{\Vert \mathbf{e}^- - \mathbf{x} \Vert - w^-(\mathbf{x})}
\end{equation}
\item \textbf{Spatial priors} refer to the 1st- and 2nd-order spatial smoothness regularizers in our objective function (17). To disable those, we set $\lambda_{1st} = \lambda_{2nd} = 0$.
\end{itemize}

\noindent
\begin{table*}[h!]
\centering
\begin{tabular}{lccccccc}
\toprule
& occlusion & coplanarity  & nonlinear $b^-$ & spatial & & \\
& reasoning & refinement   & initialization & priors & EPE rigid & EPE all \\
\midrule
\textbf{Baseline} & $\square$ & $\square$ & $\square$  &  $\square$ & 1.859 & 3.798 \\
\textbf{+occlusions} & $\blacksquare$ & $\square$ & $\square$  &  $\square$ & 1.733 & 3.705 \\
\textbf{+coplanarity} & $\blacksquare$ & $\blacksquare$ & $\square$  &  $\square$ & 1.695 & 3.671 \\
\textbf{+nonlin-init} & $\blacksquare$ & $\blacksquare$ & $\blacksquare$  &  $\square$ & 1.619 & 3.628 \\
\textbf{Full} & $\blacksquare$ & $\blacksquare$ & $\blacksquare$  &  $\blacksquare$ & 1.602 & 3.614 \\
\bottomrule
\end{tabular}
\caption{Errors when successively switching on parts of the algorithm}
\label{tab:ablation}
\end{table*}

Table~\ref{tab:ablation} shows the improvement when successively switching on more parts of the algorithm.
The occlusion reasoning has the largest positive impact on the error, since it allows the algorithm to properly merge the flow in both directions from the reference frame.
Following this, the most important parts are the nonlinear initialization and ensuring the coplanarity.
The spatial priors serve mostly to remove flow noise near boundaries. 
This improves the result visually, but has a fairly small numerical impact.

\begin{table*}[h!]
\centering
\begin{tabular}{lccccccc}
\toprule
& occlusion & coplanarity  & nonlinear $b^-$ & spatial & & \\
& reasoning & refinement   & initialization & priors & EPE rigid & EPE all \\
\midrule
\textbf{no-occlusions} & $\square$ & $\blacksquare$ & $\blacksquare$ &  $\blacksquare$ & 1.809 & 3.759 \\
\textbf{no-coplanarity} & $\blacksquare$ & $\square$ & $\blacksquare$ &  $\blacksquare$ & 1.642 & 3.645 \\
\textbf{no-nonlin-init} & $\blacksquare$ & $\blacksquare$ & $\square$ &  $\blacksquare$ & 1.677 & 3.656 \\
\textbf{no-spatial-priors} & $\blacksquare$ & $\blacksquare$ & $\blacksquare$ &  $\square$ & 1.619 & 3.628 \\
\textbf{Full} & $\blacksquare$ & $\blacksquare$ & $\blacksquare$  &  $\blacksquare$ & \textbf{1.602} & \textbf{3.614} \\
\bottomrule
\end{tabular}
\caption{Errors when disabling individual parts of the algorithm}
\label{tab:disabling}
\end{table*}

Table~\ref{tab:disabling} shows the impact when turning off individual components, but leaving all others intact.
Again, we can observe that disabling the occlusion reasoning has the largest negative impact, followed by the nonlinear initialization and ensuring the coplanarity.

\begin{table*}[h!]
\centering
\begin{tabular}{lccccc}
\toprule
& & 1st order & 2nd order & & \\
& Data term & regularization & regularization & EPE rigid & EPE all \\
\midrule
\textbf{No-opt} & $\square$ & $\square$ & $\square$ & 1.6124 & 3.6214 \\
\textbf{No-spatial-priors} & $\blacksquare$ & $\square$ & $\square$ & 1.6194 & 3.6285 \\
\textbf{No-1st} & $\blacksquare$ & $\square$ & $\blacksquare$ & 1.6025 & 3.6138 \\
\textbf{No-2nd} & $\blacksquare$ & $\blacksquare$ & $\square$ & 1.6183 & 3.6274 \\
\textbf{Full} & $\blacksquare$ & $\blacksquare$ & $\blacksquare$ & 1.6024 & 3.6138 \\
\bottomrule
\end{tabular}
\caption{Influence of regularization terms}
\label{tab:regularization}
\end{table*}

Table~\ref{tab:regularization} shows the impact of the different terms of the variational refinement (Eq.~(17)).
The cases are as described above.
In addition, table~\ref{tab:regularization} includes a complete omission of the variational refinement (\textit{no-opt}), which uses only the intial structure estimate as described by Eq.~(9) in the paper, and selective disabling of the individual regularizers ($\lambda_{1st}=0$ for \textit{no-1st} and $\lambda_{2nd}=0$ for \textit{no-2nd}).

When using the data term only (\textit{no-spatial-priors}), the error is higher than when not using any optimization.
Due to effects in the Sintel final pass such as motion blur, fog, vignetting etc. this is to be expected.
Using the 2nd order regularization improves the results; interestingly, however, the impact of the 1st order regularization is negligible.

\section{Failure cases}
\newcommand{\failurewidth}{0.3\textwidth}
This section gives examples of when our algorithm fails.
We consider our algorithm to fail when the flow computed by our algorithm is worse than the initial flow~\cite{Menze2015GCPR}.
For each of Sintel clean, Sintel final, and KITTI we give one failure example.
All examples are among the worst overall in the respective training sets.
In our training set, we observe two primary sources of error, segmentation failures and alignment failures.

\begin{figure*}[ht]
\captionsetup{justification=centering}
\centering
\begin{subfigure}[t]{\failurewidth}
	\includegraphics[width=\textwidth]{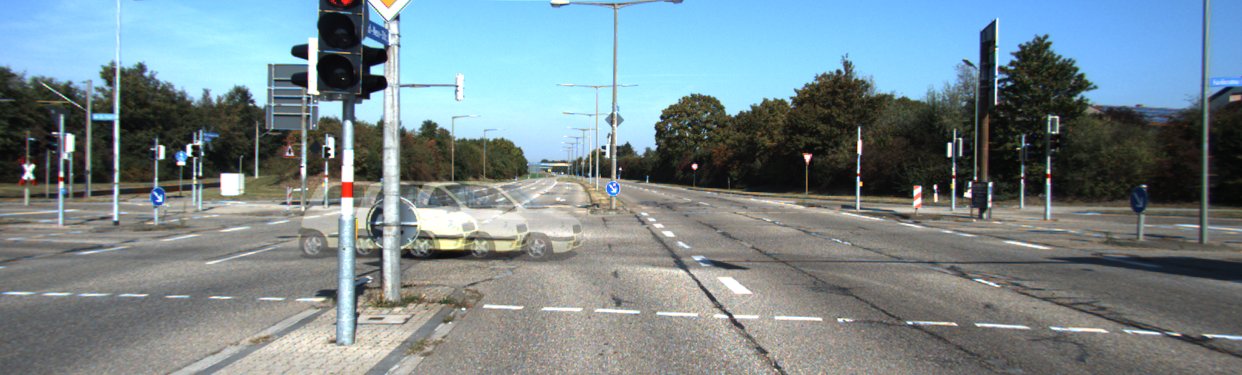}
	\caption{Input images}
\end{subfigure}%
\begin{subfigure}[t]{\failurewidth}
	\includegraphics[width=\textwidth]{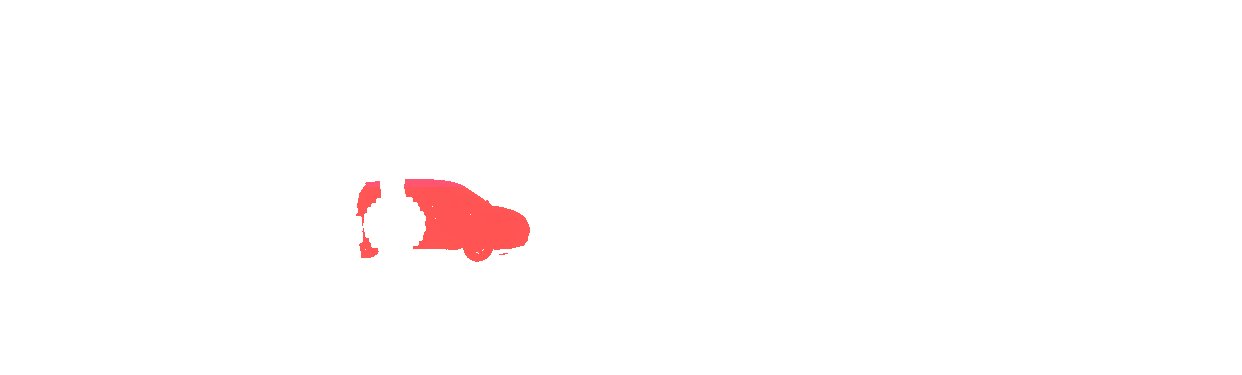}
	\caption{Ground truth flow}
\end{subfigure}%
\begin{subfigure}[t]{\failurewidth}
	\includegraphics[width=\textwidth]{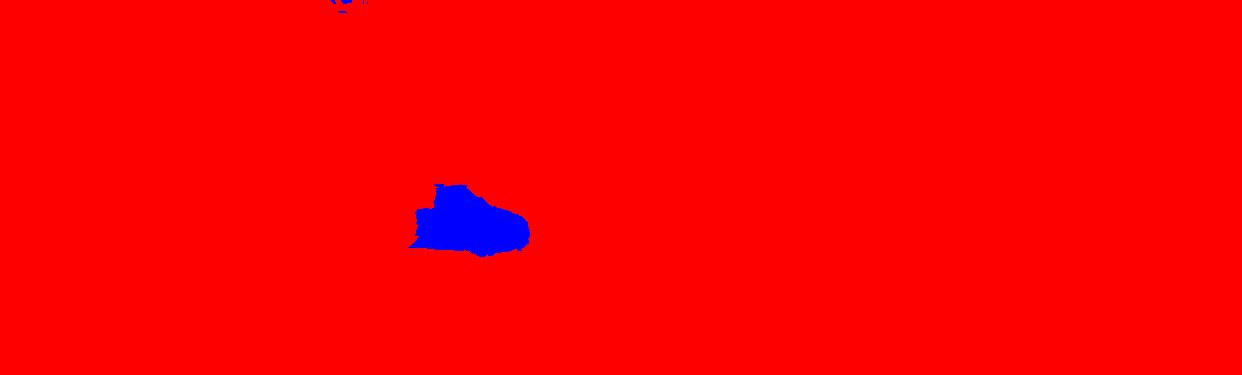}
	\caption{Estimated rigid regions}
\end{subfigure}
\begin{subfigure}[t]{\failurewidth}
	\includegraphics[width=\textwidth]{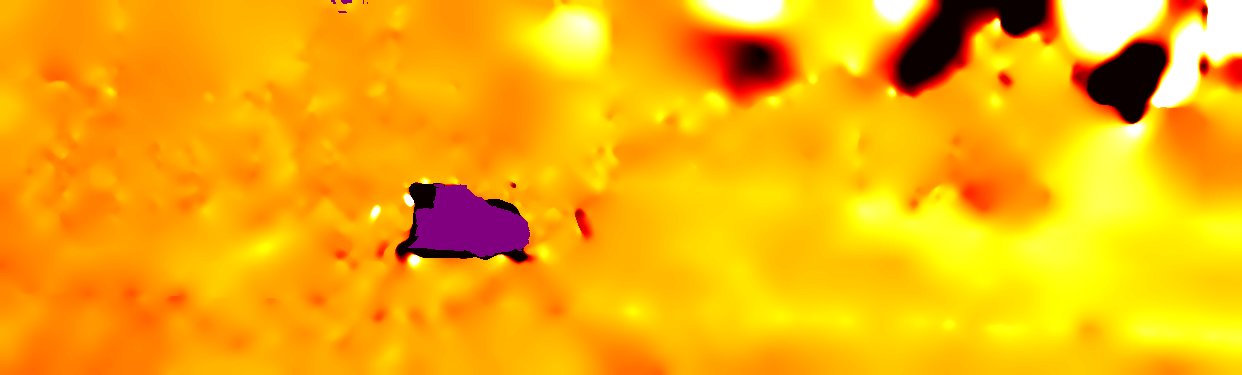}
	\caption{Estimated structure}
\end{subfigure}%
\begin{subfigure}[t]{\failurewidth}
	\includegraphics[width=\textwidth]{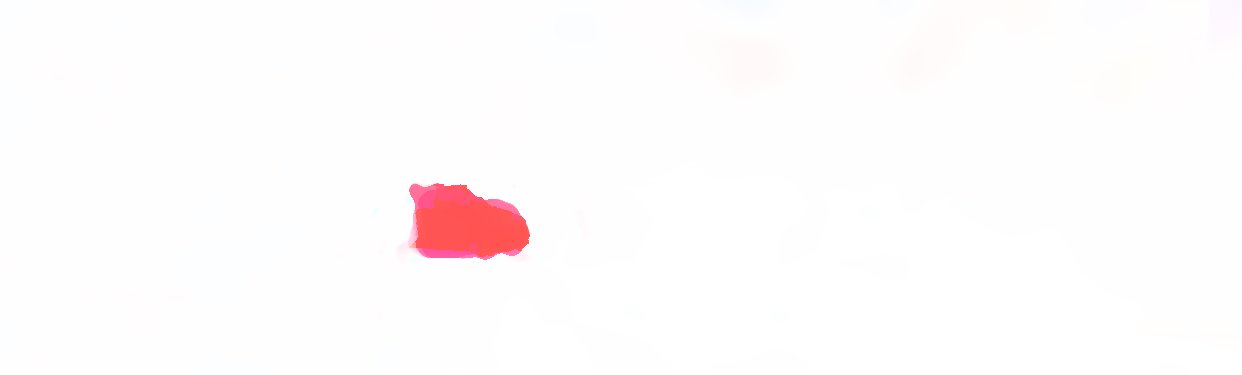}
	\caption{Estimated flow}
\end{subfigure}%
\begin{subfigure}[t]{\failurewidth}
	\includegraphics[width=\textwidth]{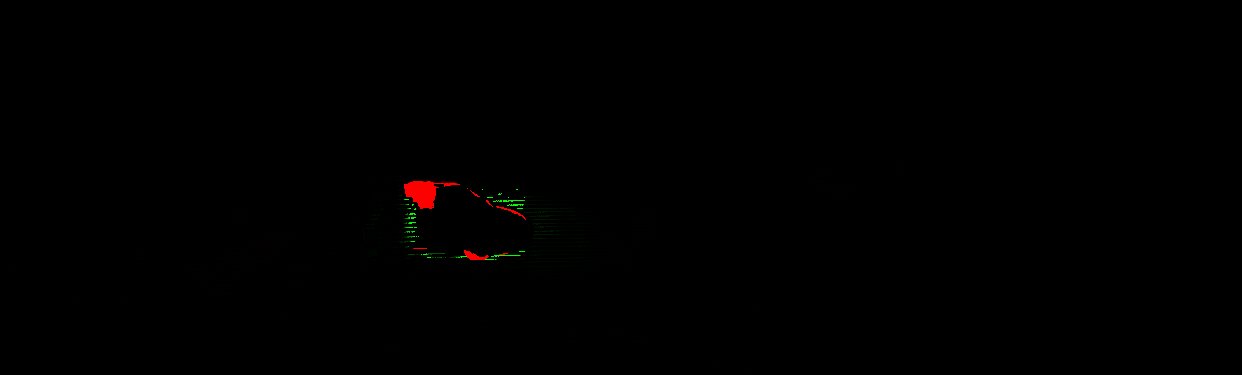}
	\caption{Comparison to initial flow}
\end{subfigure}
\caption{Failure case KITTI: The car on the left is wrongly detected as rigid. \\ Perc. wrong initialization: 3.62\%. Perc. wrong MR-Flow: 4.48\%. }
\label{fig:failure_kitti_clean}
\end{figure*}

\begin{figure*}[ht!]
\captionsetup{justification=centering}
\centering
\begin{subfigure}[t]{\failurewidth}
	\includegraphics[width=\textwidth]{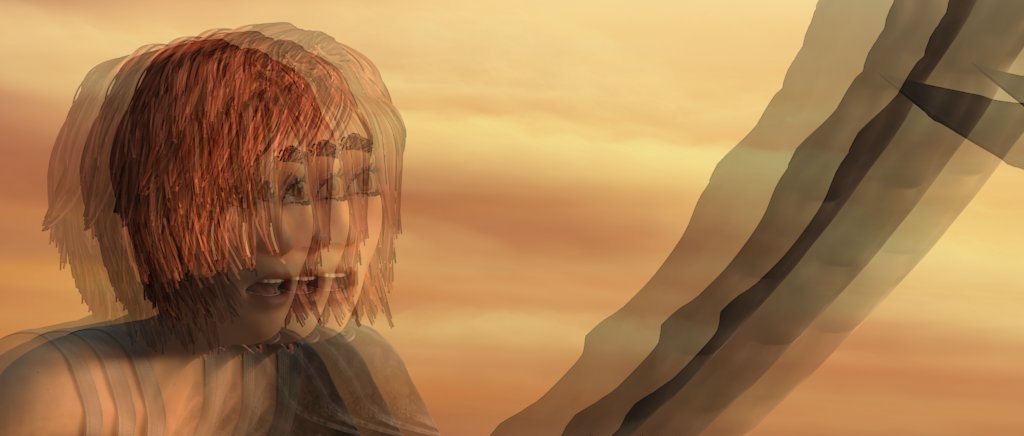}
	\caption{Input images}
\end{subfigure}%
\begin{subfigure}[t]{\failurewidth}
	\includegraphics[width=\textwidth]{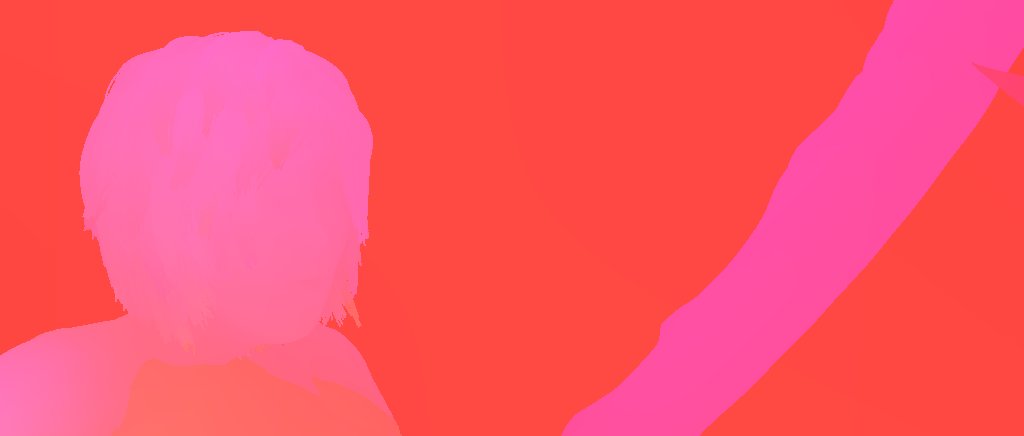}
	\caption{Ground truth flow}
\end{subfigure}%
\begin{subfigure}[t]{\failurewidth}
	\includegraphics[width=\textwidth]{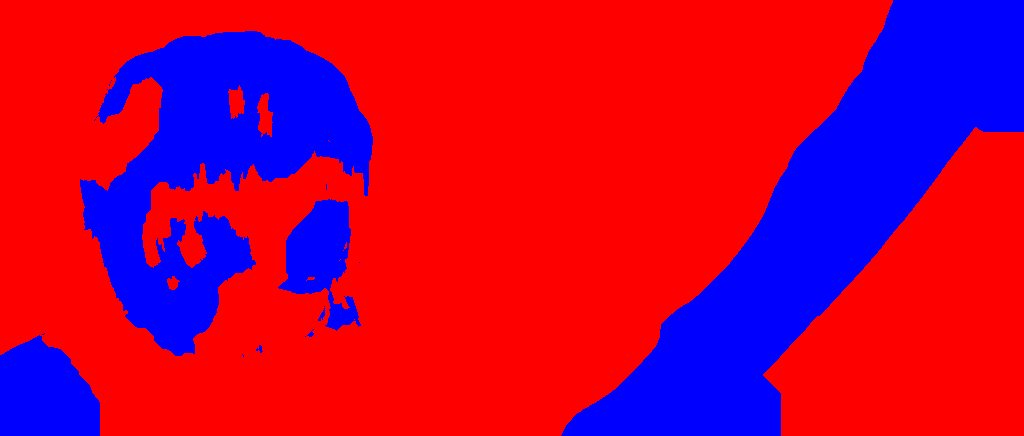}
	\caption{Estimated rigid regions}
\end{subfigure}
\begin{subfigure}[t]{\failurewidth}
	\includegraphics[width=\textwidth]{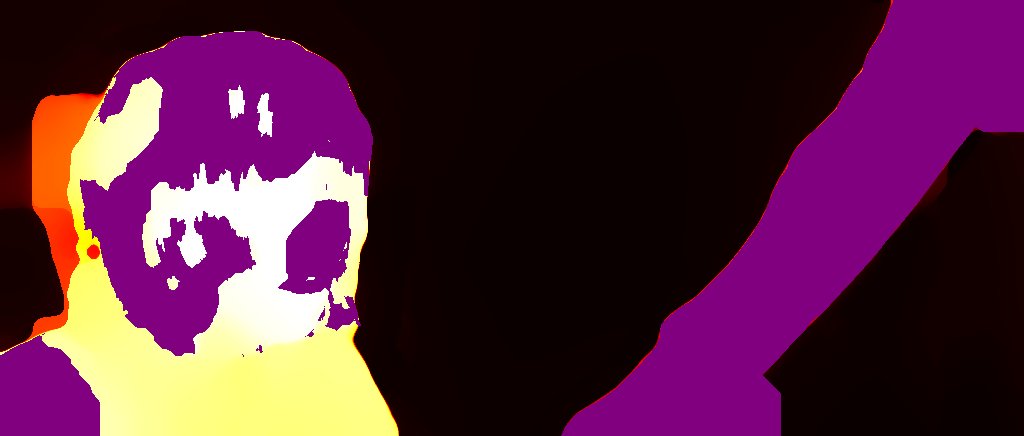}
	\caption{Estimated structure}
\end{subfigure}%
\begin{subfigure}[t]{\failurewidth}
	\includegraphics[width=\textwidth]{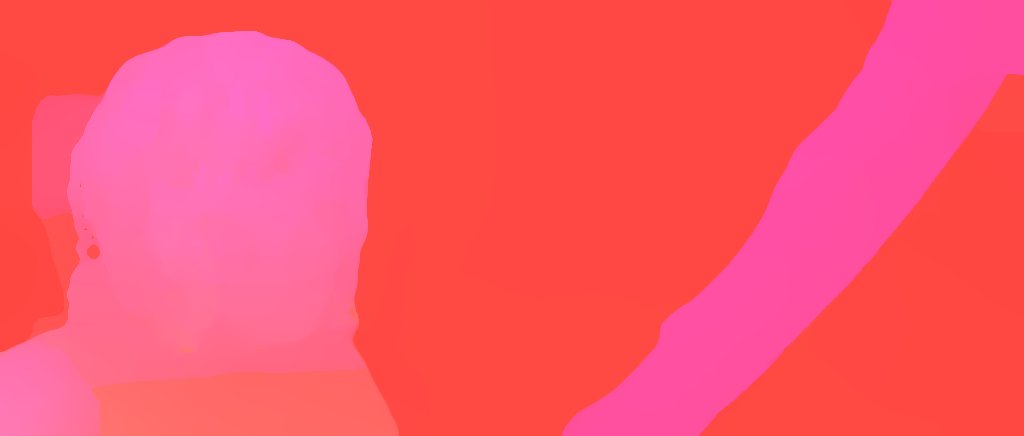}
	\caption{Estimated flow}
\end{subfigure}%
\begin{subfigure}[t]{\failurewidth}
	\includegraphics[width=\textwidth]{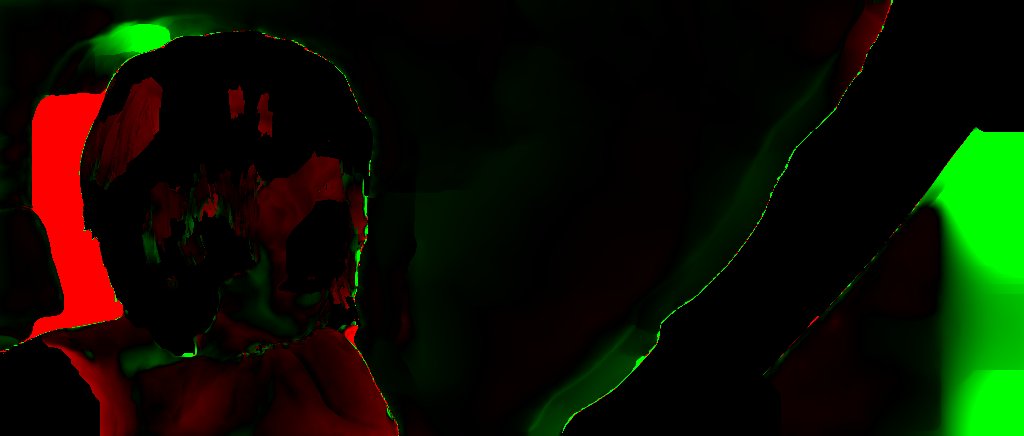}
	\caption{Comparison to initial flow}
\end{subfigure}
\caption{Failure case Sintel clean: Moving regions are wrongly detected as rigid. \\EPE initialization: $0.93$. EPE MR-Flow: $1.59$.}
\label{fig:failure_sintel_clean}
\end{figure*}

Figures~\ref{fig:failure_kitti_clean} and Figure~\ref{fig:failure_sintel_clean} show examples for the first type of error, segmentation failures.
In these cases, moving regions are mistaken as parts of the rigid background, such as the car in Fig.~\ref{fig:failure_kitti_clean} or the girl's head in Fig.~\ref{fig:failure_sintel_clean}.
These failures occur if the CNN does not pick up a region strongly enough and if, at the same time, the motion of the object is consistent with the motion of the rigid scene.
In Fig.~\ref{fig:failure_kitti_clean}, the CNN picks up only the frontal part of the car.
Since in this example the camera is not moving, the focus of expansion is mistakenly determined by the few parts of the frame that move (\ie the car), and the motion-based rigidity estimation cannot correct the mistake made by the CNN.

In Fig.~\ref{fig:failure_sintel_clean}, the camera pans to the left, and at the same time, the head moves to the right. Since both directions are approximately parallel, the head is considered to be rigid.
Note how in this case the estimated flow has a very similar hue to the ground truth flow, even in most of the regions that are misclassified.
This confirms that the direction of the flow is approximately consistent with the motion of the rigid parts of the scene; however, since the head still moves slightly out of the rigidity constraints, our method increases the error over the initialization.

\begin{figure*}[h!]
\captionsetup{justification=centering}
\centering
\begin{subfigure}[t]{\failurewidth}
	\includegraphics[width=\textwidth]{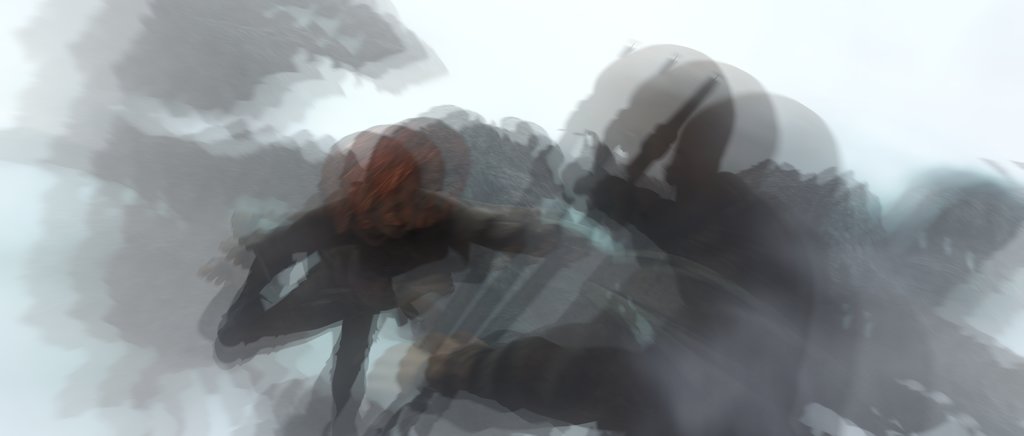}
	\caption{Input images}
\end{subfigure}%
\begin{subfigure}[t]{\failurewidth}
	\includegraphics[width=\textwidth]{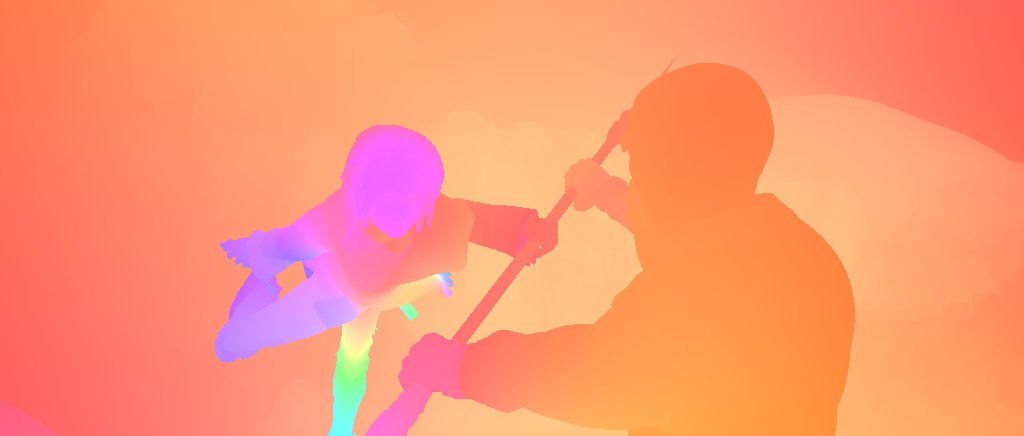}
	\caption{Ground truth flow}
\end{subfigure}%
\begin{subfigure}[t]{\failurewidth}
	\includegraphics[width=\textwidth]{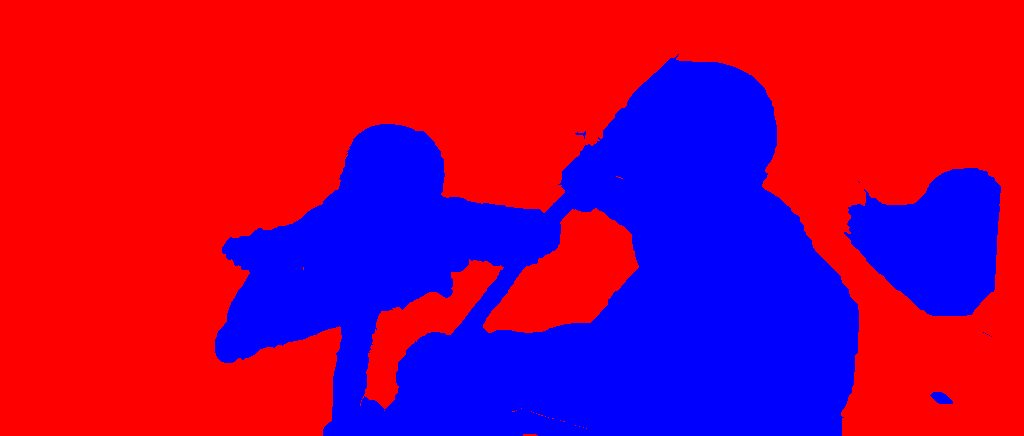}
	\caption{Estimated rigid regions}
\end{subfigure}
\begin{subfigure}[t]{\failurewidth}
	\includegraphics[width=\textwidth]{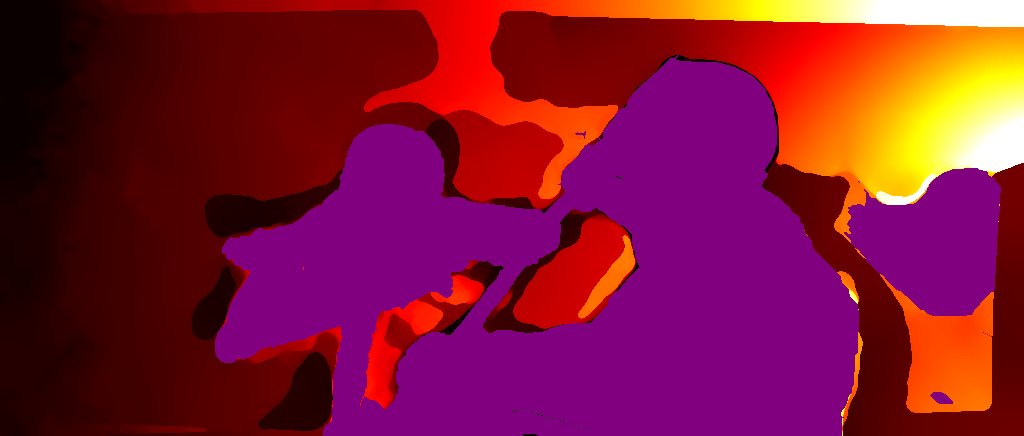}
	\caption{Estimated structure}
\end{subfigure}%
\begin{subfigure}[t]{\failurewidth}
	\includegraphics[width=\textwidth]{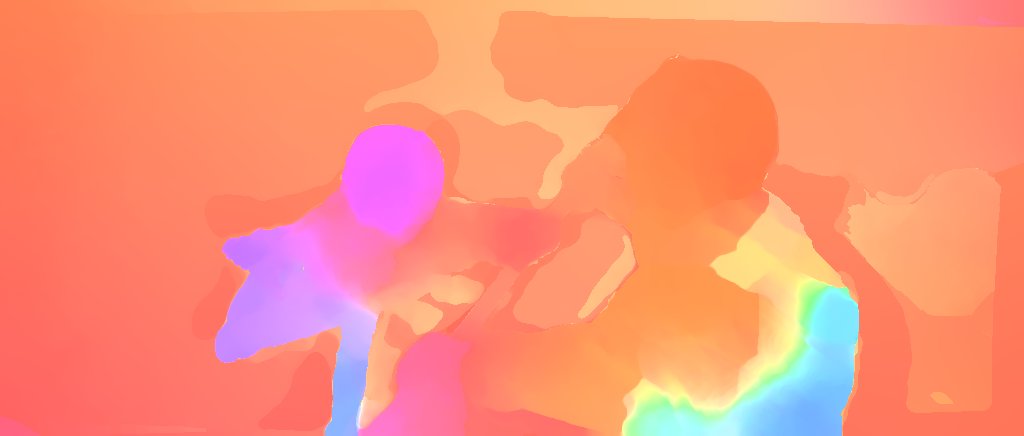}
	\caption{Estimated flow}
\end{subfigure}%
\begin{subfigure}[t]{\failurewidth}
	\includegraphics[width=\textwidth]{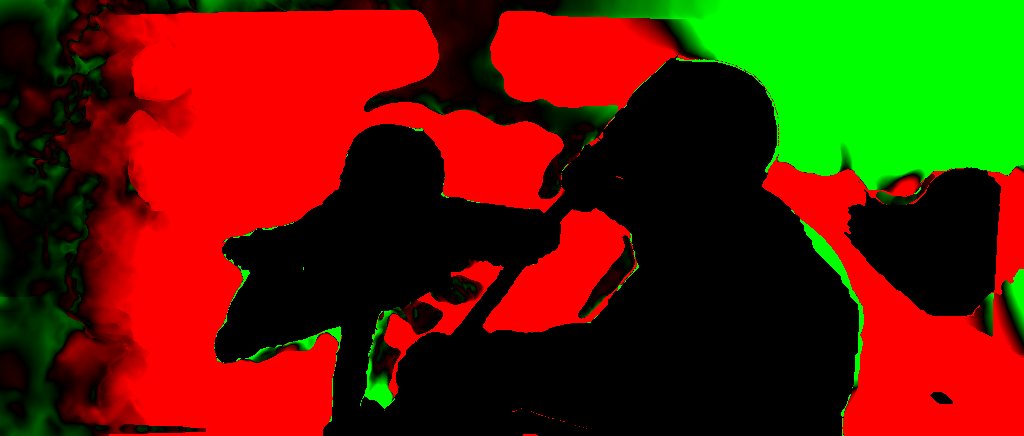}
	\caption{Comparison to initial flow}
\end{subfigure}
\caption{Failure case Sintel final: Strong motion blur destroys the alignment. \\EPE initialization: $11.17$. EPE MR-Flow: $12.21$.}
\label{fig:failure_sintel_final}
\end{figure*}

Figure~\ref{fig:failure_sintel_final} shows the second type of error, a failure to align the images.
As can be seen in Fig.~\ref{fig:failure_sintel_final}(a), the background in this sequence contains heavy motion blur and a slight vignetting.
Together, these two effects cause a high uncertainty of the initial optical flow in the background regions, which in turn causes our initial alignment procedure to fail.

\end{document}